\documentclass{article}

\usepackage[preprint]{neurips_2020}
\usepackage{graphicx} 
\usepackage{caption,subcaption}
\usepackage{float}

\usepackage[utf8]{inputenc} 
\usepackage[T1]{fontenc}    

\usepackage{booktabs}       
\usepackage{amsfonts}       
\usepackage{nicefrac}       
\usepackage{microtype}      
\usepackage{multirow}

\usepackage{comment}
\usepackage{amsmath}
\usepackage{graphicx}

\usepackage{authblk}

\usepackage{hyperref}       
\usepackage{url}            

\title{MACE: Model Agnostic Concept Extractor for Explaining Image Classification Networks}

\author{Ashish Kumar \thanks{2016csb1033@iitrpr.ac.in}}
\author{Karan Sehgal \thanks{2016csb1080@iitrpr.ac.in}}
\author{Prerna Garg \thanks{2016csb1050@iitrpr.ac.in}}
\author{Vidhya Kamakshi \thanks{2017csz0005@iitrpr.ac.in}}
\author{Narayanan C Krishnan \thanks{ckn@iitrpr.ac.in}}
\affil{IIT Ropar}

\begin{document}
\maketitle

\begin{abstract}
Deep convolutional networks have been quite successful at various image classification tasks. The current methods to explain the predictions of a pre-trained model rely on gradient information, often resulting in saliency maps that focus on the foreground object as a whole. However, humans typically reason by dissecting an image and pointing out the presence of smaller concepts.  The final output is often an aggregation of the presence or absence of these smaller concepts.  In this work, we propose MACE: a Model Agnostic Concept Extractor, which can explain the working of a convolutional network through smaller concepts.  The MACE framework dissects the feature maps generated by a convolution network for an image to extract concept based prototypical explanations. Further, it estimates the relevance of the extracted concepts to the pre-trained model's predictions, a critical aspect required for explaining the individual class predictions, missing in existing approaches.  We validate our framework using VGG16 and ResNet50 CNN architectures, and on datasets like Animals With Attributes 2 (AWA2) and Places365. Our experiments demonstrate that the concepts extracted by the MACE framework increase the human interpretability of the explanations, and are faithful to the underlying pre-trained black-box model. 


\end{abstract}

\section{Introduction}
\begin{figure}[H]
\centering
\includegraphics[width=\linewidth, height=\linewidth, keepaspectratio]{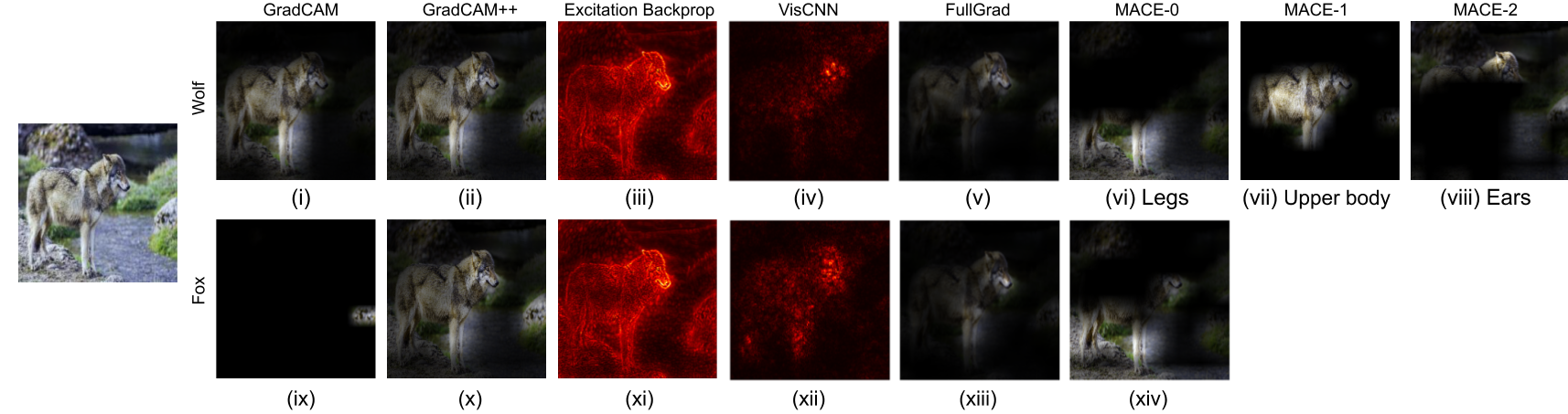}
\caption{[Best viewed in color] Explanations generated by different approaches for the predicted class and another class for an image of wolf.}
\label{fig:comparison}
\end{figure}

The state of the art convolutional networks has been quite successful at various computer vision tasks. However, the improved performance has come at the cost of reduced human understanding of the model. It is crucial to bring transparency in these networks to help understand their decisions. Recently, there has been a lot of work on designing models that explain the behavior of a pre-trained convolutional network. These approaches generate saliency maps for explaining the model's behavior or explain the prediction based on training instances. Approaches such as GradCAM \cite{gradCAM} and its variants, use gradient information to generate saliency maps as illustrated in Figure \ref{fig:comparison}. Other approaches like Excitation Backpropagation \cite{DBLP:journals/corr/Zhang0BSS16} use top-down neural attention to generate attention maps. These explanations provide a high-level insight into the working of the model. However, we observe that invariably almost the entire object is highlighted in all the saliency maps and their explanations are almost identical for the predicted class and any other class for a given image. This reduces the interpretability of their explanations. While the explanations can detect the foreground object, they are unable to accurately highlight regions in the object that contributed towards the prediction. 

\begin{figure}[H]
    \centering 
\begin{subfigure}{0.49\textwidth}
  \includegraphics[width=0.24\linewidth]{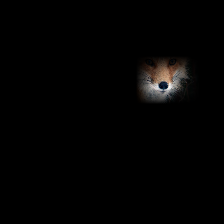}%
  \hfill
  \includegraphics[width=0.24\linewidth]{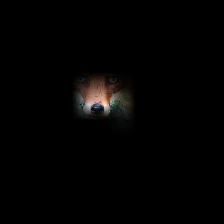}%
    \hfill
  \includegraphics[width=0.24\linewidth]{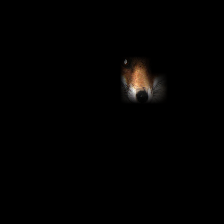}%
    \hfill
  \includegraphics[width=0.24\linewidth]{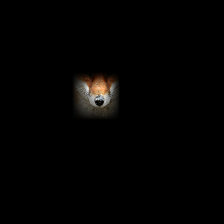}
  \caption{Muzzle in class Fox.}
  \label{fig:muzzle}
\end{subfigure}
\hfill
\begin{subfigure}{0.49\textwidth}
  \includegraphics[width=0.24\linewidth]{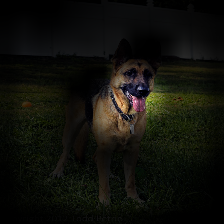}%
  \hfill
  \includegraphics[width=0.24\linewidth]{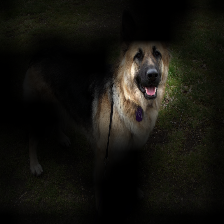}%
    \hfill
  \includegraphics[width=0.24\linewidth]{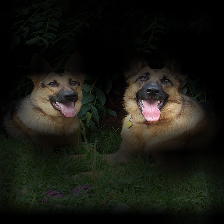}%
    \hfill
  \includegraphics[width=0.24\linewidth]{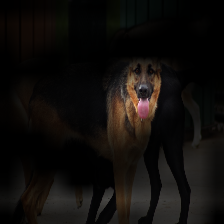}
  \caption{Face in class German Shepherd.}
  \label{fig:face}
\end{subfigure}
\\
\begin{subfigure}{0.49\textwidth}
  \includegraphics[width=0.24\linewidth]{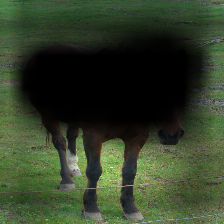}%
  \hfill
  \includegraphics[width=0.24\linewidth]{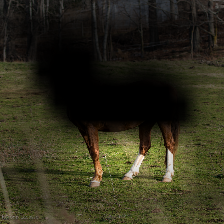}%
    \hfill
  \includegraphics[width=0.24\linewidth]{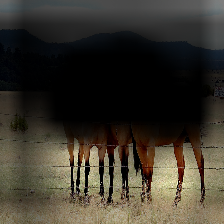}%
    \hfill
  \includegraphics[width=0.24\linewidth]{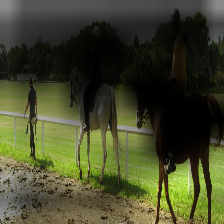}
  \caption{Legs in class Horse.}
  \label{fig:legs}
\end{subfigure}
\hfill
\begin{subfigure}{0.49\textwidth}
  \includegraphics[width=0.24\linewidth]{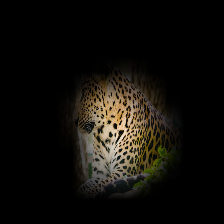}%
  \hfill
  \includegraphics[width=0.24\linewidth]{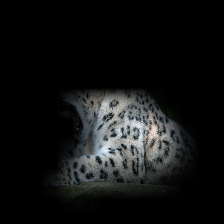}%
    \hfill
  \includegraphics[width=0.24\linewidth]{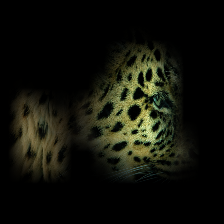}%
    \hfill
  \includegraphics[width=0.24\linewidth]{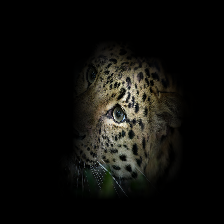}
  \caption{Body texture in class Leopard.}
  \label{fig:body}
\end{subfigure}
\caption{[Best viewed in color] Visualization of concepts across multiple images of the same class.}
\label{fig:outputIllustration}
\end{figure}

Another style of reasoning for a prediction is by dissecting an image and referring to the presence of smaller concepts. For example, if the image is that of a lion, the smaller concepts could be legs, ears, face, skin texture, etc. The aggregation of these smaller concepts is used to explain the final output. The fundamental objective of our framework: Model-Agnostic Concept Extractor (MACE) is to mimic this style of reasoning; which is to explain the model's behavior in terms of smaller concepts in the image. Our approach learns multiple concepts for a class without any supervision or additional information. These concepts are visualized through multiple high quality localized saliency maps, rather than a single region. Figure \ref{fig:outputIllustration} shows the visualizations of some of the concepts learned by our approach.

Further, the proposed framework also estimates the relevance of each concept with respect to the output of the model. This is particularly helpful when the explanation for any output of the model is desired. In Figure \ref{fig:comparison} the first row corresponds to the explanations from different approaches for the predicted class and the second row presents the explanations for the class with second-highest prediction probability. It can be noticed that there is no significant difference in the explanations generated by current approaches for the two outputs of the model. However, our approach can generate class-specific concepts as well as their relevance for a more comprehensive explanation. For an incorrect class, most of the concepts have negative relevance thereby suppressing the prediction probability. As illustrated in Figure \ref{fig:comparison} while the predicted class is the wolf, the second-highest probability is for class fox. The MACE framework extracts only a single concept (that of legs) with positive relevance for the class fox.

We also ensure that the MACE framework generates explanations that satisfy a few desirable properties. \textit{Stability}: The learned concepts should be consistent i.e. the localized saliency maps generated for a particular concept should be similar across all the images of the same class, thus achieving higher interpretability. This can be observed in Figure \ref{fig:outputIllustration} where the visualizations for a single concept obtained from different images appear very similar. \textit{Faithfulness}: The explanations generated by the framework should accurately represent the working of the pre-trained model. This will improve the trustworthiness of the explanations. Finally, \textit{Robustness}: The explanations should be robust to perturbations such as noise, translation, and rotation in the input. Explanations of perturbed inputs should not change significantly if there is no significant change in the corresponding outputs.
\section{Related Work}
The last few years have seen significant work on developing post-hoc explanation models \cite{gradCAM, grad_cam_plus_plus, cam, lime, anchors, shap}. Some of the popular approaches construct saliency maps that highlight important regions in the image. The Class Activation Map (CAM) \cite{cam} is one of the earliest approaches to construct a saliency map specific to a classification model. The gradient-class activation maps (Grad-CAM) \cite{gradCAM} is a generalization of CAM that uses gradients flowing into the final convolutional layer to localize salient image regions. GradCAM is among the most popular approach for generating explanations for an image classification network. However, it has multiple drawbacks including poor localization in the presence of multiple instances of the same object in an image and poor distinction of the class-specific saliency maps generated for different classes, given an input \cite{similar_saliency_ieee}. GradCAM++ \cite{grad_cam_plus_plus}, a variant of GradCAM, overcomes the limitation of generating accurate saliency maps for multiple instances of an object in an image. However, GradCAM++ requires the estimation of higher-order derivatives, the cost of which grows with increasing network complexity. 

Vis-CNN \cite{vis_cnn} also uses gradients to explain the contribution of each input pixel towards the output. But pixel-wise quantification appears like a foreground detector looking at most of the results presented in the paper and also during our experiments. Full Grad \cite{full_gradient} combines the idea of using gradients flowing back to intermediary layers as well as gradients flowing back till input to generate a saliency map attributing importance of each pixel towards a prediction. On the other hand, Wang et al \cite{similar_saliency_ieee} suggest a possible compromise in the faithfulness of explanations when gradients are used to generate the explanations. There have been other variants of CAM like Ablation-CAM \cite{ablation_cam} and Score-CAM \cite{score_cam} that avoid the need for gradients for computing the saliency maps. Our approach also estimates saliency maps for explaining the pre-trained model without utilizing the gradient information. Further, we automatically synthesize multiple maps each characterizing a different concept present in the image.

Fong et al, \cite{perturbation_deletion} propose a perturbation based approach for estimating the salient regions in the image, that was further extended to also take into account when the region was preserved \cite{perturbation_preservation}. These complementary approaches aim to modify the input image successively and map the region integral for the final prediction. There are many challenges with the feasibility of the approach given the high dimensional nature of the images and voluminous amount of possible perturbations. Excitation backpropagation \cite{DBLP:journals/corr/Zhang0BSS16} proposes a probabilistic attention mechanism to explain the working of black boxes in a fine-grained manner. But as recent works like \cite{attention_mohankumar_2020,attention_ieee_xu,grimsley2020attention} state, attention need not be human interpretable .


Another category of explainable models is the ante-hoc models that are explainable by design \cite{DLFCBP,this_looks_like_that,hierarchical_prototypes,concept_whitening}. Due to explainability being a parameter considered in the design, the explanations are faithful to the underlying model. However, the need to retrain the model from scratch to incorporate the explainability aspect is an obstacle to explain a pre-trained model that has already been deployed. 

Our proposed approach is a posthoc explainability technique like \cite{cam,gradCAM,grad_cam_plus_plus} that leverages some aspects from ante-hoc explainability techniques like \cite{this_looks_like_that, DLFCBP} to generate fine-grained and faithful explanations. Like GradCAM, our approach generates human interpretable explanations in the form of a saliency map. However, unlike GradCAM, we do not utilize the gradient backflow information for generating the saliency maps. Our work is closely related to the exciting paradigm of `this looks like that' proposed by Chen et al \cite{this_looks_like_that}. Specifically, the MACE framework generates multiple saliency maps containing different parts/concepts for explaining the output of a single image. However, unlike \cite{this_looks_like_that}, our approach does not require modifying and retraining the black box architecture nor is there a need to slide across the activation maps or tune additional parameters to visualize a concept.

\section{Methodology}
The MACE framework is envisioned as a modular network that can be attached to any convolution layer of a pre-trained network for probing its functionality. However, for describing the framework, we assume that the MACE network is inserted in between the final convolutional layer and the first dense layer of a pre-trained network. This lateral connection taps into the spatial distribution of concepts (as gathered from the downstream convolutional layers) and the discriminative features for classification (from the upstream dense layers). Given, a query image and a class label, the MACE framework extracts the class-specific low-level concepts along with their relevance towards the output of the model. 

\begin{figure}
    \centering
    \includegraphics[width=\textwidth,height=\textheight,keepaspectratio]{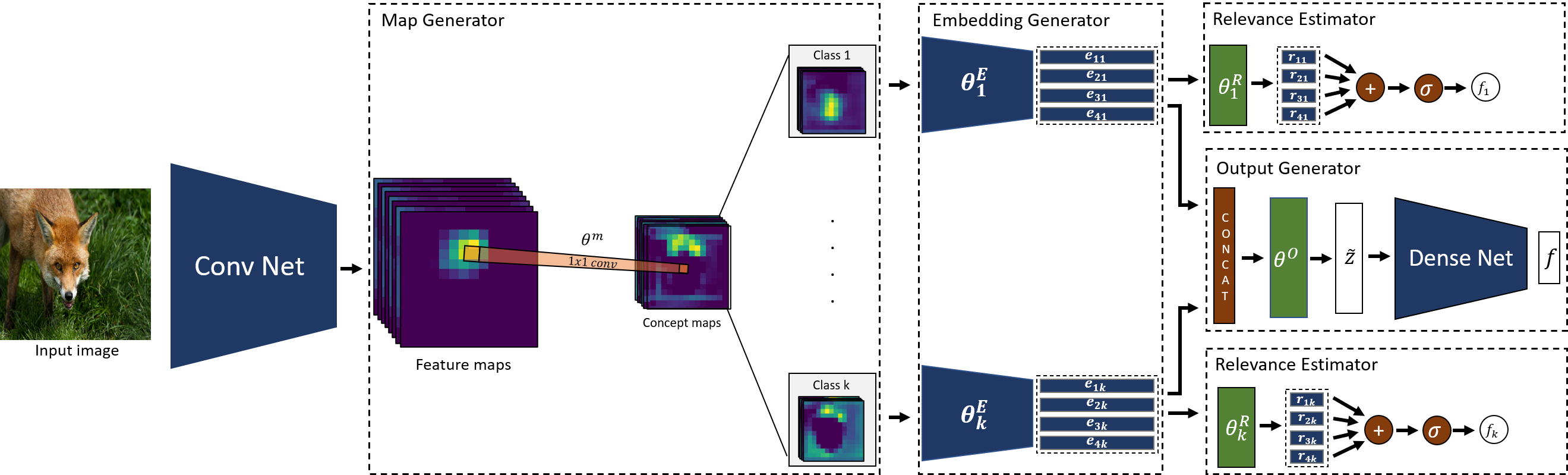}
    \caption{[Best viewed in color] Architecture of the MACE Framework.}
    \label{fig:architecture}
\end{figure}

The process of learning and extracting the concepts and their relevance is divided across four modules namely; the map generator, embedding generator, relevance estimator, and output generator. The MACE framework is illustrated in Figure \ref{fig:architecture}. The map generator learns an attended spatial map representing the spread of a concept using the output of the last convolutional layer of the pre-trained model. The embedding generator transforms the manifestation of a concept into a latent representation that is invariant to spatial information. The relevance estimator determines the importance of a class-specific concept towards the output of the model. Finally, the output generator completes the loop by linking the extracted concepts back to the first dense layer of the pre-trained model.

Let $f$ be a pre-trained model for a $K$-way image classification task; trained using a dataset $\mathcal{D}$. The output of the last convolutional layer in $f$ for a given input is denoted as $\textbf{x}$. Thus $\textbf{x}\in \mathbb{R}^{H\times W\times D}$, where $H$ and $W$ refer to the size of the feature map and $D$ refers to the number of filters in the last convolution operation. Let $\textbf{z}\in \mathbb{R}^L$ represent the output of the dense layer following the convolutional layer. The MACE framework is laterally connected in between the transformation of $\textbf{x}$ to $\textbf{z}$. 

For the sake of simplicity, we assume that every class can be explained by the same number of concepts. (In practice as discussed in the supplementary material section S2, pruning may result in a varying number of concepts per class). A concept has two aspects to it. The first is its actual manifestation in an image, termed as the concept map, for example, the region corresponding to the ear or legs. The second aspect is the latent representation, termed as the concept embedding, that is invariant to the manifestation.  Irrespective of the location and orientation of the concept ear in an image, the encoding of the concept remains the same. $\textbf{c}_{jk} \in \mathbb{R}^{H\times W}$ denotes the $j^{th}$ concept map for the $k^{th}$ class and $\textbf{e}_{jk} \in \mathbb{R}^Q $ denotes the corresponding concept embedding.

\subsection{Map Generator}
The map generator takes as input the convolutional feature map $\textbf{x}$ and estimates a concept map $\textbf{c}_{jk}$. The concept map encodes two properties - the activation pattern of a concept and the salient region in the image that expresses the activation pattern. The activation pattern is used by the embedding generator to extract a distinct and invariant encoding for the concept. The salient region in the image is used for visualizing the concept post-training, facilitating the human interpretability of the concept.

Due to the nature of convolution operations, the activations of a concept are distributed across the different channels of the feature map $\textbf{x}$. The MACE framework combines the channels in the feature map to produce a single activation pattern for a concept. This is achieved as a weighted average of the channels in the feature map. Specifically, we employ 1D convolutions to obtain the concept map. Let $\theta_{jk}^M$ denote the weights of the 1D convolution filter for generating the $j^{th}$ concept map of the $k^{th}$ class, then $  \textbf{c}_{jk} = ReLU(\textbf{x} * \theta_{jk}^M) $. The $ReLU$ operation is performed because, the locations in the activation map corresponding to large positive values may denote the presence of the concept \cite{gradCAM,grad_cam_plus_plus}. We have to learn $K\times C$ 1D convolution filters, assuming $C$ concepts per class. The concept maps $\textbf{c}_{jk}$ are of the same size as the input feature map $\textbf{x}$. 

Each of these concept maps represents an activation pattern. The region in the concept map exhibiting the highest activation is most likely to have expressed the concept. This region in the original image contributing to this activation can be visualized by resizing the concept map to the original image dimension. We use this resized concept map to visualize the concept. The MACE framework does not impose any constraints on the size of the concept. Thus allowing it to learn small concepts such as ears to large concepts like body and background. 


\subsection{Embedding Generator}
The next module - the embedding generator takes as input the concept maps that are stacked class-wise. The map generator only uses the convolution layer activation maps to generate the concept map and thus retains the spatial information about the location of the concept in the image. Consider two images one in which the object lion is present on the right half of the image, and a different orientation of the object lion is present in the left half of the second image. The concept maps of these two images will have high activations in different regions even though the underlying concept is the same. Thus, we need to abstract out the concept from its actual manifestation. We refer to this encoding as the concept embedding. We would like the embeddings for a concept extracted from different images belonging to the same class exhibiting the concept to form a tight cluster.

The abstraction of the concept through the embedding also enables a comparison of different concepts. We model the embedding generator as a multi-layer dense network. Having a sufficiently large network enables the extraction of different concepts. However, the parameters of the network are shared across the concepts of a class to reduce the overall learning complexity. Formally, let the embedding generator network for class $k$, be parameterized by $\theta^E_k$. A stack of the concept maps $\{\textbf{c}_{jk}\}_{j=1}^C$ extracted from an input image are fed as inputs to embedding generator resulting in the set of embeddings $\{\textbf{e}_{jk}\}_{j=1}^C$. Restricting to only a positive subspace (as was the case with the concept map generation) is not necessary for learning the concept embedding. We use tanh activations enabling the use of the full space to learn well-separated embeddings.

Triplet loss is used to learn the invariant concept embeddings. We normalize the embeddings before applying the triplet loss. We use the training images belonging to class $k$  from a mini-batch to learn the $C$ concept embeddings for the class. The embeddings generated for concept $j$ across the training images of a particular class are chosen as anchor positives. The embeddings for the other concepts (except $j$) for a specific training image are the anchor negatives. Similar to \cite{schroff2015facenet}, we use all anchor-positive pairs and select semi-hard negatives for anchor-negative pairs. The margin $\alpha$ is set to 1 to make the embeddings orthogonal to each other. Let $\textbf{e}_{jk}^a(i)$ represent an anchor for the $i^{th}$ image in the mini-batch, the anchor positives are represented as $\textbf{e}_{jk}^p(i)$ and the anchor negatives as $\textbf{e}_{jk}^n(i)$. Then the triplet loss for class $k$ using a mini-batch of $B$ images is defined as
\begin{equation}
\mathcal{L}_k^E = \sum_{i=1}^B\sum_{j = 1}^{C}\bigg[ \| \textbf{e}_{jk}^{a}(i) - \textbf{e}_{jk}^{p}(i)\|_{2}^{2} - \|\textbf{e}_{jk}^{a}(i) - \textbf{e}_{jk}^{n}(i)\|_{2}^{2} + \alpha\bigg]_{+}
\label{eq:tripletloss}
\end{equation}
\subsection{Relevance Estimator}
A key aspect of the MACE framework is the estimation of the relevance of the concept towards the model prediction.  This score, termed as concept relevance, quantifies the contribution of a concept towards the output of the model for a given class. The concept relevance would enable a comparison of different concepts for a given class and provide better insight into the relationship between the concepts and the model's output.

Given a training image, the concept embeddings of class $k$ are concatenated and passed through a sigmoid activated dense layer (parameterized by $\theta^R_k$) with a single output to learn the probability of classification for class $k$ as estimated by the pre-trained model. $\theta^R_k$ can be divided into chunks representing the connection weights for the individual concepts, i.e. $\theta^R_k=[\theta^R_{1k}, \ldots, \theta^R_{Ck}]$.  Then, the concept relevance for concept $j$ with respect to the output for class $k$ is denoted as $r_{jk}={\theta^R_{jk}}^T\textbf{e}_{jk}$. Note that the concept relevance $r_{jk} \in (-\infty, \infty)$. As the argument to the sigmoid activation of the dense layer is $\sum_{j=1}^Cr_{jk}$, the impact of the concept relevance scores on the prediction probabilities can be easily understood. Positive (negative) concept relevance increases (decreases) the prediction probability, thus emphasizing (reducing) the importance of the concept for a particular prediction. 

The parameters of the dense layer, $\theta^R_k$, are learned by minimizing the cross-entropy between the output of the sigmoid and the pre-trained model's prediction probability for class $k$. Specifically, for a mini-batch of training instances, the loss for learning the parameters is defined as 
\begin{equation}
    \mathcal{L}_k^R=-\sum_{i=1}^B f_k(i) \log\left(\sigma\left(\sum_{j=1}^Cr_{jk}(i)\right)\right)
\end{equation}
where $f_k(i)$ and $r_{jk}(i)$ are the pre-trained models output for class $k$ and the concept relevance for concept $j$ with respect to class $k$ respectively for the training instance $i$ .
\subsection{Output Generator}
To increase the faithfulness of the explanations we loop back the concept embeddings into the pre-trained model. The concatenated concept embeddings are passed through a dense layer (parameterized by $\theta^O$) to approximate the output, $\textbf{z}$, of the first dense layer of the pre-trained model. An $L_{2}$ loss between the approximation, $\tilde{\textbf{z}}$, and $\textbf{z}$, defined as $\mathcal{L}^D = \| \textbf{z} - \tilde{\textbf{z}}\|^{2}_{2}$ is used to learn the approximation. If the approximation is accurate we should obtain the output $f(\textbf{z})$ ($\textbf{x}$ is replaced by $\textbf{z}$, a slight abuse of the notation), when
$\tilde{\textbf{z}}$ is passed through the dense layers of the pre-trained model. This requirement is enforced by minimizing the divergence between the pre-trained model's outputs for $\tilde{\textbf{z}}$ and $\textbf{z}$ defined as $\mathcal{L}^O = KL( f(\tilde{\textbf{z}})  \|  f(\textbf{z}) )$. The $L_{2}$ loss is required to faithfully mimic the behavior of the pre-trained model. On the other hand, leaving out the divergence loss can still lead to a good approximation $\tilde{z}$, but $f(\tilde{z})$ can be inconsistent with $f(z)$.

Overall, the MACE framework minimizes $\mathcal{L}=\sum_{k=1}^K(\mathcal{L}^E_k + \mathcal{L}^R_k) + \mathcal{L}^D + \mathcal{L}^O$ with respect to the parameters $\theta^M, \theta^E, \theta^R$, and $\theta^O$. Adam optimizer \cite{adam} is used to minimize the loss function.

\section{Experiments}
We validate the MACE framework on VGG \cite{simonyan2014deep}, and ResNet \cite{DBLP:journals/corr/HeZRS15} architectures trained on the Imagenet dataset \cite{DBLP:journals/corr/RussakovskyDSKSMHKKBBF14}. The networks are fine-tuned on AWA2 dataset \cite{AWA2,awa2_paper_xian_2018_zero} and the Places365 dataset \cite{places}. We discuss the results for the AWA2 dataset using the VGG model in the paper and present the results for the Places365 dataset and ResNet architecture in the supplementary material (sections S8 and S7 respectively). The AWA2 dataset consists of 37322 images of 50 animal classes. We select a subset of 10 classes with approximately 400 images per class for our experiments. The VGG network that is fine-tuned with this dataset is our pre-trained model. For every class, we aim to learn 10 concepts ($C=10$) with the concept embedding dimension to be 32 ($Q=32$). Examples of the concepts extracted for various classes are presented in Figure \ref{fig:outputIllustration}. The architectural details of the embedding generator, along with the choices for the hyper-parameters are presented in the supplementary material section S1. Additional examples of the visualization of the concepts for the AWA2 dataset are presented in sections S3 and S4 of the supplementary material. While we describe the results from three salient experiments here, investigations into the stability and robustness of the explanations as well as ablation studies are discussed in sections S5, S6, and S9 of the supplementary material respectively. 

\subsection{Faithfulness of the Explanations}
\begin{figure}[htb]
    \centering 
\begin{subfigure}{0.495\textwidth}
  \includegraphics[width=\linewidth]{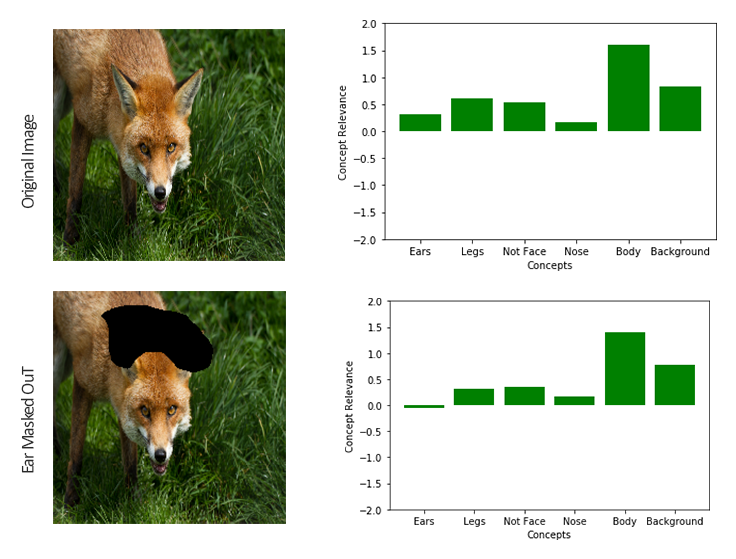}
  \caption{}
  \label{fig:ear_masked}
\end{subfigure}\hfil 
\begin{subfigure}{0.495\textwidth}
  \includegraphics[width=\linewidth]{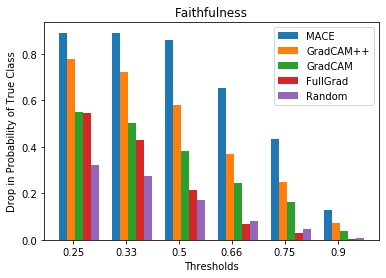}
  \caption{}
  \label{fig:faithfulness}
\end{subfigure}\hfil 
\caption{[Best viewed in color] Faithfulness of the explanations extracted by MACE (a) No change in the prediction distribution after masking a concept due to compensation by the model using other concepts, (b) Drop in the  probability of classification across different threshold values for generating the binary mask.}
\label{fig:faithfullness_full}
\end{figure}

Faithfulness, the ability to accurately explain the working of a pre-trained model, is one of the most important traits of a post-hoc explanation model. As suggested in prior literature, we measure faithfulness using the drop in the classification probability upon masking. We generate a binary mask by thresholding each of the concept maps. As the concepts are trained to be different from each other, masking out just one concept might not give a significant drop in the probability of the predicted class, as the model can still predict with the help of other concepts. For example, as seen in Figure \ref{fig:ear_masked} masking the ear concept does not alter the classification probabilities significantly as the model compensates for it through the other concepts. Thus, we take the union of all binary masks. The region in the generated mask is removed from the image and the resulting drop in the probability of the predicted class is measured.  We also compare the drop in the prediction probability when the mask is synthesized using other saliency-map generating approaches as well as a random process (activation maps combined randomly). The results of this experiment are presented in Figure \ref{fig:faithfulness}. The highest drop in the probability is observed for the mask synthesized by the MACE framework for every threshold.
\subsection{Explaining all the outputs for a test image}
\begin{figure}[htb]
    \centering 
\begin{subfigure}{0.70\textwidth}
\centering
  \includegraphics[width=\linewidth]{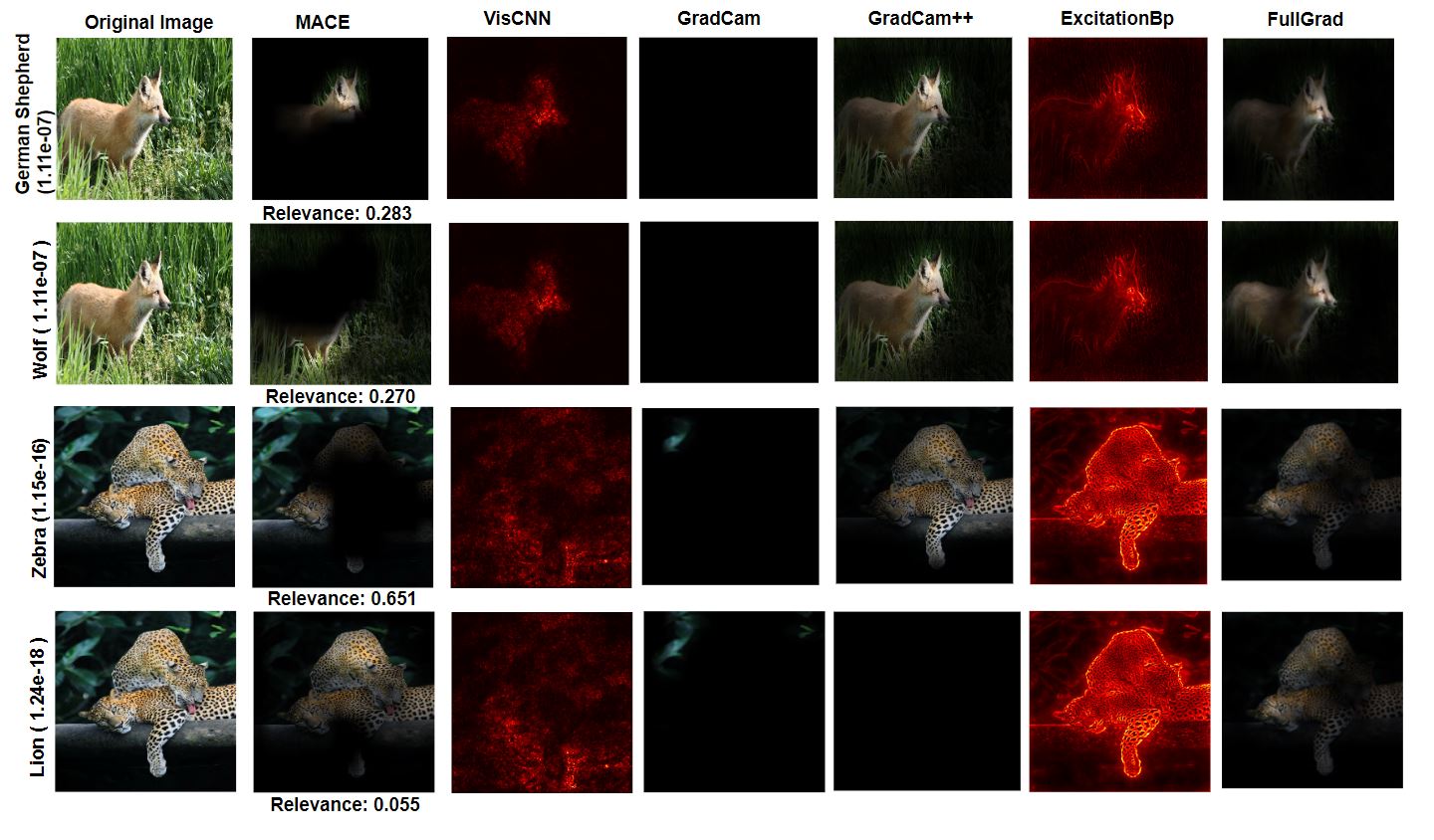}
  \caption{}
  \label{fig:wrongexp}
\end{subfigure}\hfil 
\begin{subfigure}{0.29\textwidth}
\centering
  \includegraphics[width=0.95\linewidth, keepaspectratio]{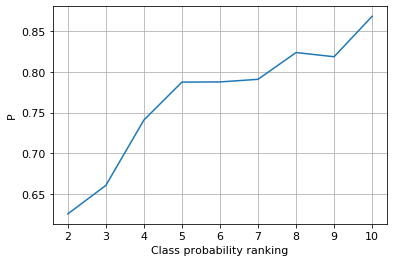}
  \caption{}
  \label{fig:pvp}
  \includegraphics[width=0.95\linewidth, keepaspectratio]{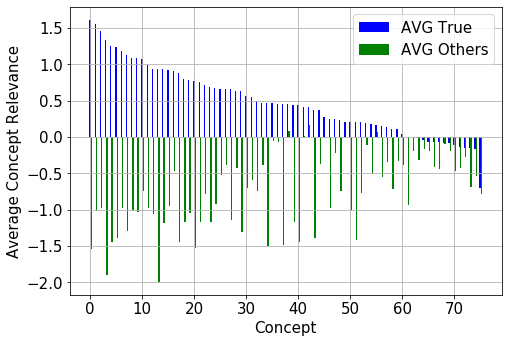}
  \caption{}
  \label{fig:bargraph}
\end{subfigure}\hfil 
\caption{[Best viewed in color] (a) contains explanations of different approaches for incorrect class. Each row corresponds to the explanation for some output probability for a wrong class. The wrong class is written on the left-most side in each row. The image shown for MACE approach is the visualization of the concept with the highest concept relevance for that class, and the concept relevance is written below the visualization; (b) Percentage of concepts with positive concept relevance less than the average score using only the images for different classes (See text for more details); (c) Average relevances of all the concepts for the true class and for all the classes except the true class.}
\label{fig:faithfullness_full}
\end{figure}

An image classification network outputs a probability distribution over the class labels.  An important aspect of any explanation model is to explain all the non-zero probabilities in the output of the classifier. In our approach, this is explained by looking at the concepts generated for every class, and the corresponding concept relevances. We specifically look at the concepts that have positive concept relevance for any class. Figure \ref{fig:wrongexp} shows the results for a few images. For example, the first two rows present the explanations generated by different models for why the classifier has non-zero probabilities for the German Shepherd and Wolf classes when the correctly predicted class is Fox. The original image in the first column of \ref{fig:wrongexp} is a fox image, the visualization of explanation given by MACE (second column) for German Shepherd class and wolf class are meaningful and highlight smaller regions in the fox such as ears and legs, while other approaches give the same explanation for both the classes. We can also observe from the figure that MACE consistently gives different meaningful explanations for different classes, while other approaches give the same or no explanation. 

  
We further analyze the behavior of the concept relevance scores for different outputs of the pre-trained model. We first compute the average concept relevance for a concept embedding $e_{jk}$ using all the images belonging to class $k$. Over the test set, we compute the percentage of concepts with positive concept relevance less than the average score using only the images for which class $k$ is the second-highest ranked class according to the pre-trained model's predictions. We compute the average percentage across all classes for a particular rank. The result is summarized in Figure \ref{fig:pvp}. We can observe from the figure that the percentage of concepts with positive relevance score less than the average increases as the prediction ranking decreases. This is intuitive as concept relevance for the concept of an incorrect class should decrease as we go down the class ranking.

Figure \ref{fig:bargraph} shows the average concept relevances for all the concepts where the average is split between relevances of concepts belonging to the true class of a test image (AVG True) and relevances of concepts belonging to any other class except the true class (AVG Others). It can be observed that AVG Others (green bars) is significantly lower than AVG True (blue bars) for all the concepts. This further strengthens the credibility of concept relevances as concept relevance should be higher only if the concept actually belongs to the true class of the image than any other class.

\subsection{Human Evaluation of Explanations}
We evaluate the quality of our explanations generated for the VGG16 network fine-tuned on 10 classes of the AWA2 dataset. We present to every user the saliency maps generated by MACE, GradCAM \cite{gradCAM}, GradCAM++ \cite{grad_cam_plus_plus}, Excitation Backpropagation \cite{DBLP:journals/corr/Zhang0BSS16} and VisCNN \cite{vis_cnn} for a random set of 10 images. In addition, we randomly select 2 images from this set for repetition and create a final questionnaire of 12 examples and the corresponding explanations. The subjects were asked to choose the approach whose explanation helped them better understand the classifier's prediction. The responses in which the answers to the repeated questions mismatched were removed to maintain the consistency in the responses. We received 41 consistent responses, resulting in a total of 410 votes. The results have been summarised in Table \ref{HumanExptable}. We observe that explanations from MACE are the most preferred choice amongst the participants. Further, approaches like MACE and Excitation Backpropagation are preferred over gradient-based approaches like VisCNN, GradCAM and GradCAM++. 
\begin{table}
\centering
\caption{Preferences of the Human Subjects for Explaining the Classifier's Predictions.}
 \begin{tabular}{||c | c||} 
 \hline
 Approach & Vote Percentage \\ [0.5ex] 
 \hline
 \hline
 MACE (ours) & \textbf{48.29\%} \\ 
 \hline
 Excitation Backpropagation \cite{DBLP:journals/corr/Zhang0BSS16} & 40.73\% \\
 \hline
 GradCAM++ \cite{grad_cam_plus_plus} & 9.51\% \\
 \hline
 GradCAM \cite{gradCAM} & 1.21\% \\
 \hline
 VisCNN \cite{vis_cnn} & 0.24\% \\
 \hline
\end{tabular}
\label{HumanExptable}
\end{table}

\section{Summary}
In this work, we define a new form of reasoning for explaining the outputs of an image classification network using multiple concepts/parts of an object. We present the MACE framework, for extracting and visualizing these multiple concepts. We also propose a mechanism for estimating the relevance of a concept towards the output of the model. We perform extensive experiments on the MACE framework using VGG16 and ResNet50 architectures for animal and places classification tasks. Our results confirm the faithfulness of the explanations as well as their human interpretability.

\newpage
\bibliographystyle{plainnat}
\bibliography{neurips2020.bib}
\end{document}


\maketitle

\section{Experimental Details}

Our implementation can be accessed at \url{https://github.com/mace19/MACE}. In this section, we discuss our experimental details for various architectures and datasets. For each dataset, we select 10 classes and for every class, we aim to learn 10 concepts with concept embedding dimension equal to 32. 

\subsection{VGG16 on AWA2}
We train our MACE framework on VGG16 \cite{simonyan2014deep} architecture using a subset of AWA2 dataset \cite{AWA2,awa2_paper_xian_2018_zero}, having 10 classes with approximately 400 images per class. The learning rate is set to $10^{-4}$ and the model is trained for 64 epochs using an ADAM optimizer \cite{adam}. The accuracy of the model was 92.7\%

\subsection{VGG16 on Places365}
We train our MACE framework on VGG16 \cite{simonyan2014deep} architecture using a subset of Places365  \cite{places}, having 10 classes with approximately 5000 images per class. The learning rate is set to $5*10^{-4}$ and the model is trained for 32 epochs using an ADAM optimizer \cite{adam}. The accuracy of the model was 93.1\%

\subsection{ResNet-50 on AWA2}
We train our MACE framework on ResNet-50\cite{DBLP:journals/corr/HeZRS15} architecture using a subset of AWA2 dataset \cite{AWA2,awa2_paper_xian_2018_zero}, having 10 classes with approximately 400 images per class. The learning rate is set to $10^{-3}$ and the model is trained for 128 epochs using an ADAM optimizer \cite{adam}. The accuracy of the model was 96.5\%


\section{Pruning}

The set of concepts generated by following the proposed methodology generate some concepts that are not meaningful. We prune these concepts so that only the meaningful concepts remain. We use the following strategy for pruning the concepts:

\begin{itemize}
\item We fetch the top $T$ images in terms of concept relevance, and if more than $S$ of those images don't belong to the same class as the concept, then we prune the concept. In our experiments, we set $T = 10$ and $S = 5$.
\item If a large part of test dataset has positive concept relevance for a concept, we prune that concept. In our experiments with AWA2 dataset, we pruned the concept if it had positive concept relevance for more than 50\% of the test images.
\item If the concept masks in almost the entire image, we prune the concept. In our case, we pruned the concept if on an average it masked in 95\% of the image.
\item If a concept has negative concept relevance for most of the images that belong to the same class as the concept, then we prune the concept. In our experiments we pruned the concept if less than 5\% of the images belonging to the same class had positive concept relevance.
\end{itemize}

\begin{figure}[H]
    \centering 
\begin{subfigure}{\textwidth}
  \includegraphics[width=0.12\linewidth]{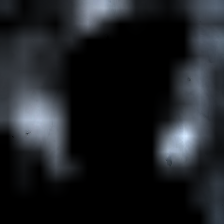}%
  \hfill
  \includegraphics[width=0.12\linewidth]{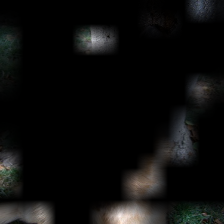}%
    \hfill
  \includegraphics[width=0.12\linewidth]{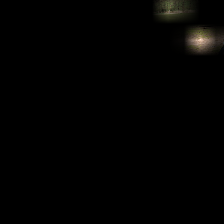}%
    \hfill
  \includegraphics[width=0.12\linewidth]{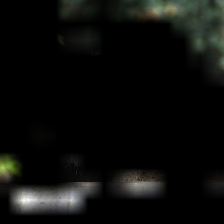}%
  \hfill
  \includegraphics[width=0.12\linewidth]{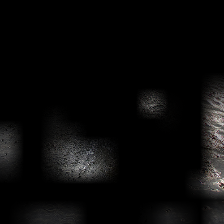}%
  \hfill
  \includegraphics[width=0.12\linewidth]{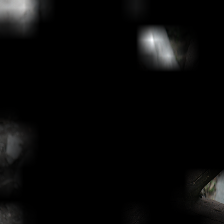}%
    \hfill
  \includegraphics[width=0.12\linewidth]{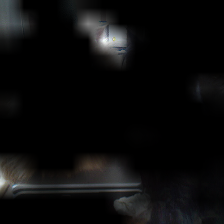}%
    \hfill
  \includegraphics[width=0.12\linewidth]{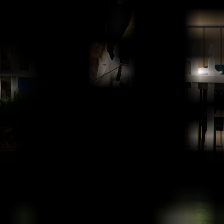}
  \label{fig:ear}
\end{subfigure}
\begin{subfigure}{\textwidth}
  \includegraphics[width=0.12\linewidth]{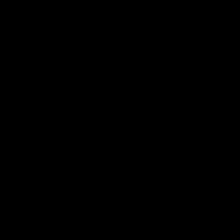}%
  \hfill
  \includegraphics[width=0.12\linewidth]{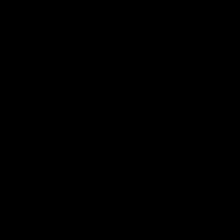}%
    \hfill
  \includegraphics[width=0.12\linewidth]{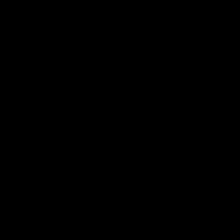}%
    \hfill
  \includegraphics[width=0.12\linewidth]{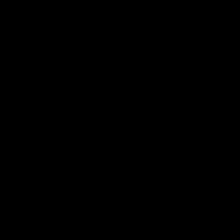}%
  \hfill
  \includegraphics[width=0.12\linewidth]{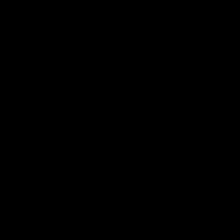}%
  \hfill
  \includegraphics[width=0.12\linewidth]{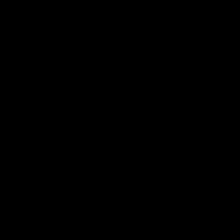}%
    \hfill
  \includegraphics[width=0.12\linewidth]{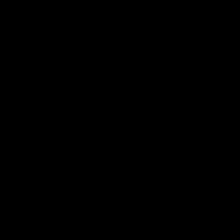}%
    \hfill
  \includegraphics[width=0.12\linewidth]{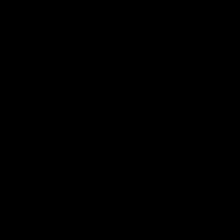}
  \label{fig:ear}
\end{subfigure}
\begin{subfigure}{\textwidth}
  \includegraphics[width=0.12\linewidth]{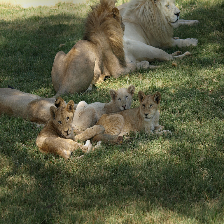}%
  \hfill
  \includegraphics[width=0.12\linewidth]{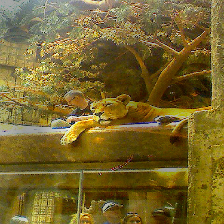}%
    \hfill
  \includegraphics[width=0.12\linewidth]{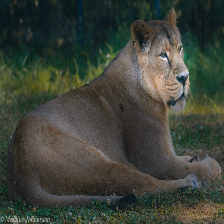}%
    \hfill
  \includegraphics[width=0.12\linewidth]{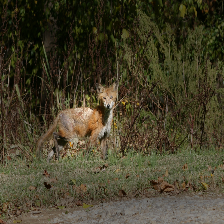}%
  \hfill
  \includegraphics[width=0.12\linewidth]{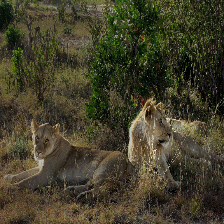}%
  \hfill
  \includegraphics[width=0.12\linewidth]{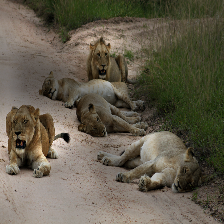}%
    \hfill
  \includegraphics[width=0.12\linewidth]{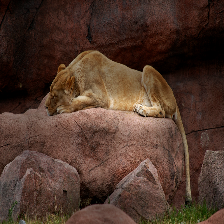}%
    \hfill
  \includegraphics[width=0.12\linewidth]{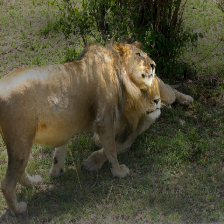}
  \label{fig:ear}
\end{subfigure}
\begin{subfigure}{\textwidth}
  \includegraphics[width=0.12\linewidth]{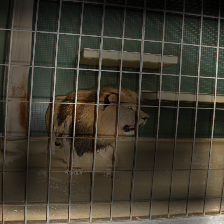}%
  \hfill
  \includegraphics[width=0.12\linewidth]{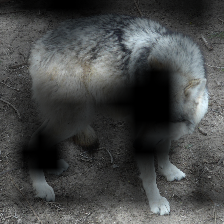}%
    \hfill
  \includegraphics[width=0.12\linewidth]{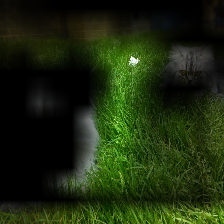}%
    \hfill
  \includegraphics[width=0.12\linewidth]{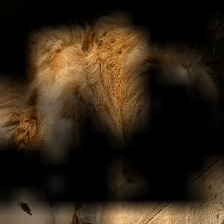}%
  \hfill
  \includegraphics[width=0.12\linewidth]{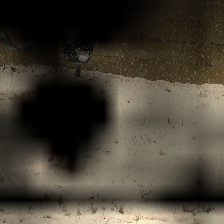}%
  \hfill
  \includegraphics[width=0.12\linewidth]{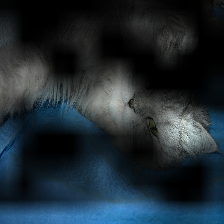}%
    \hfill
  \includegraphics[width=0.12\linewidth]{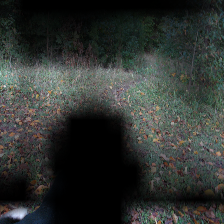}%
    \hfill
  \includegraphics[width=0.12\linewidth]{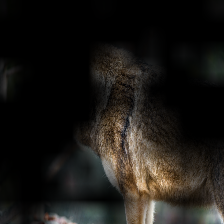}
  \label{fig:ear}
\end{subfigure}
\begin{subfigure}{\textwidth}
  \includegraphics[width=0.12\linewidth]{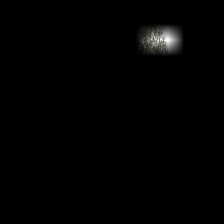}%
  \hfill
  \includegraphics[width=0.12\linewidth]{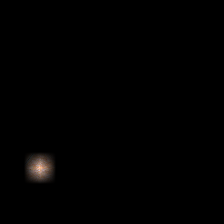}%
    \hfill
  \includegraphics[width=0.12\linewidth]{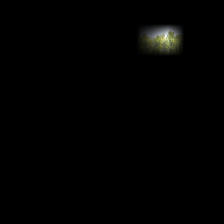}%
    \hfill
  \includegraphics[width=0.12\linewidth]{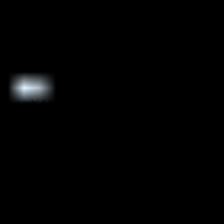}%
  \hfill
  \includegraphics[width=0.12\linewidth]{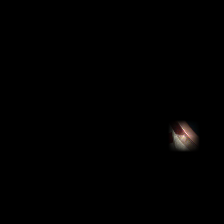}%
  \hfill
  \includegraphics[width=0.12\linewidth]{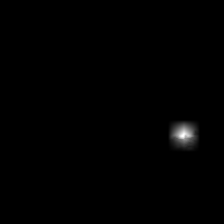}%
    \hfill
  \includegraphics[width=0.12\linewidth]{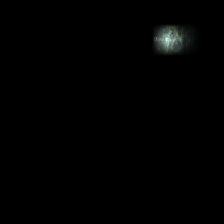}%
    \hfill
  \includegraphics[width=0.12\linewidth]{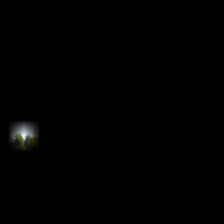}
  \label{fig:ear}
\end{subfigure}
\caption{[Best viewed in color] Some of the pruned concepts. Each row contains prototypical images of a pruned concept.}
\label{fig:pruned_concepts_visualization}
\end{figure}

After pruning the concepts, we fine-tune the MACE model. We use the fine-tuned model to perform all the experiments. Visualizations of some of the pruned concepts are shown in Figure ~\ref{fig:pruned_concepts_visualization}

\newpage
\section{Concept visualization and relevance for true class prediction}
\begin{figure}[H]
    \centering
    \includegraphics[width=\textwidth]{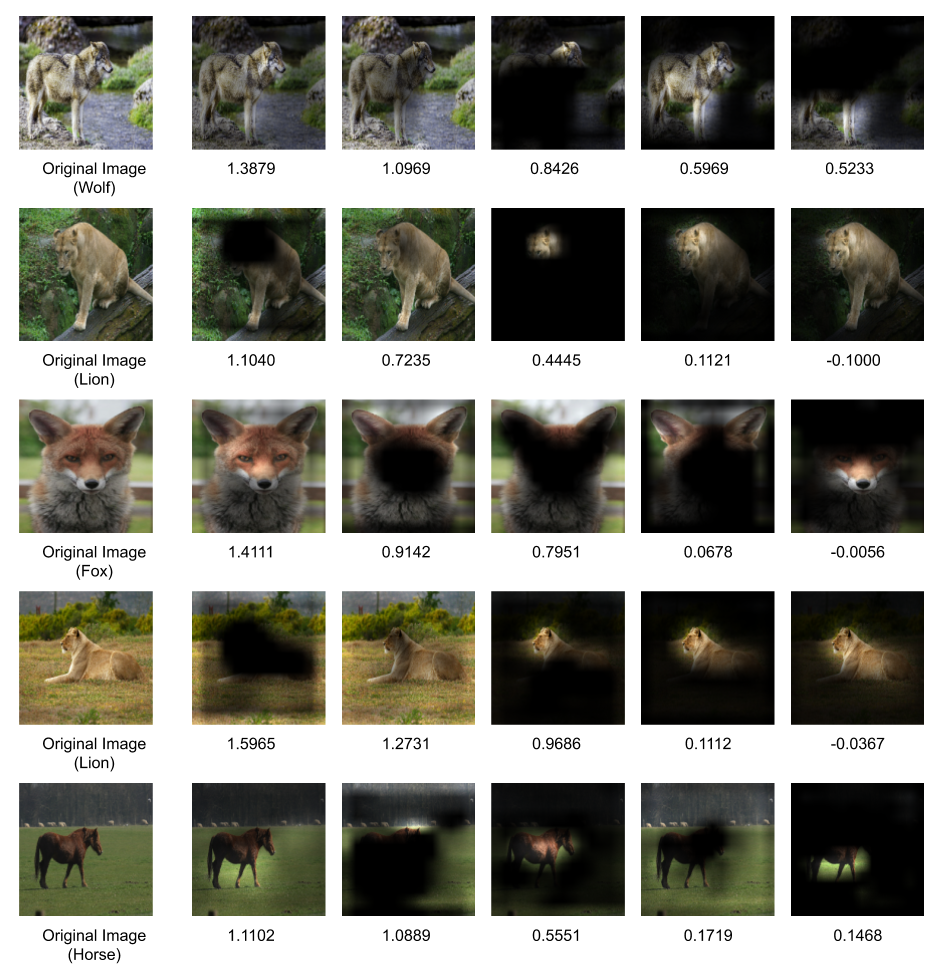}
    \caption{[Best viewed in color] Concepts with relevances corresponding to the predicted class}
    \label{fig:first_5}
\end{figure}

Figure \ref{fig:first_5} shows the visualizations of the learned concepts and their corresponding relevance values for a few images. We observe that for most images the background concept had very high relevance thus suggesting that it plays an important contribution in the model's prediction. We also observe some repetitive pattern in the visualizations of the learned concepts. Such repetitions have been removed from this figure.

\newpage
\section{Concept visualization and relevance for other class prediction}
\begin{figure}[H]
    \centering
    \includegraphics[width=\textwidth]{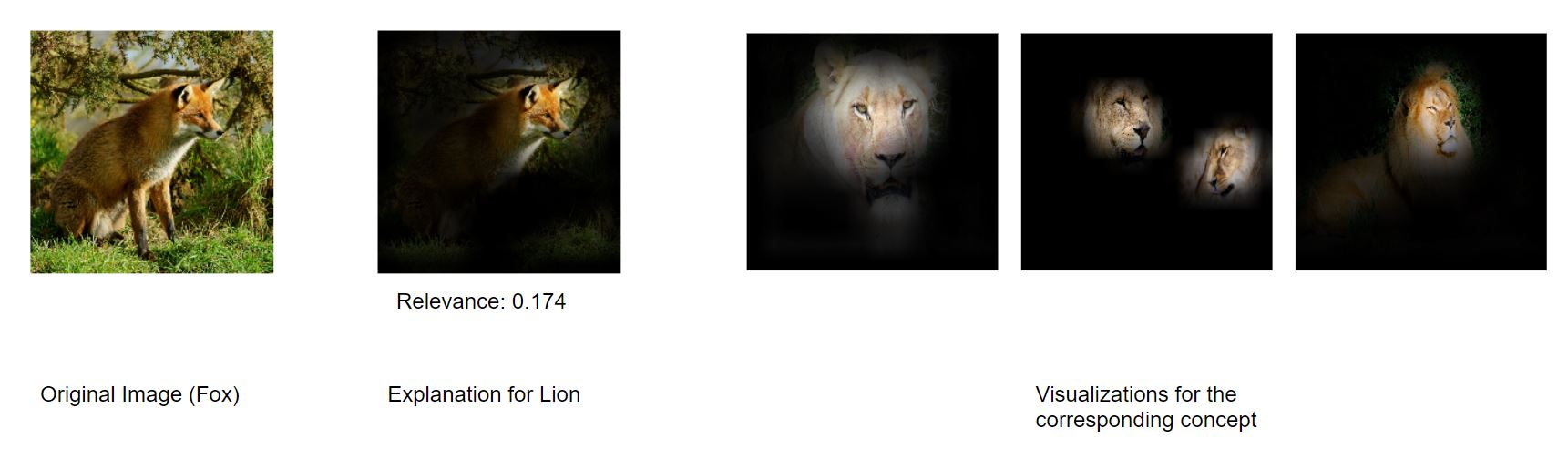}
    \caption{[Best viewed in color] The model is looking at the face region in the fox image in this concept. The visualizations for lion for this concept is also face.}
    \label{fig:fox_lion_face}
\end{figure}

\begin{figure}[H]
    \centering
    \includegraphics[width=\linewidth]{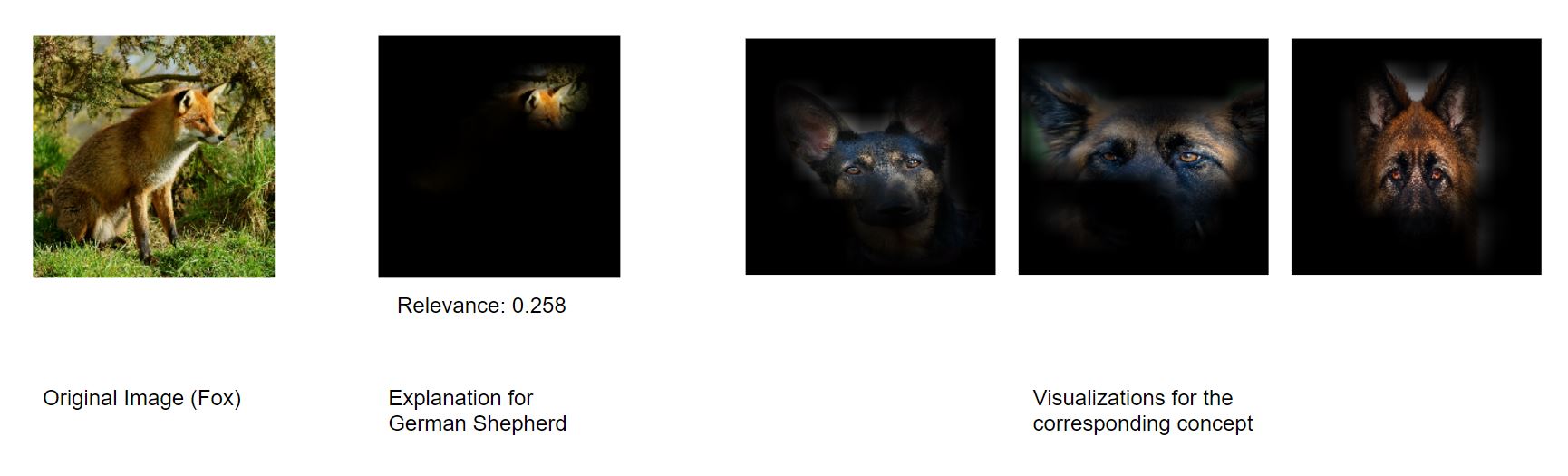}
    \caption{[Best viewed in color] The model is looking at the ear region in the fox image in this concept. The visualizations for dog for this concept is the ear and eye region. Since eyes are not clearly visible in the fox image, the model is looking at just the ears in this concept.}
    \label{fig:fox_dog_ear}
\end{figure}

\begin{figure}[H]
    \centering
    \includegraphics[width=\linewidth]{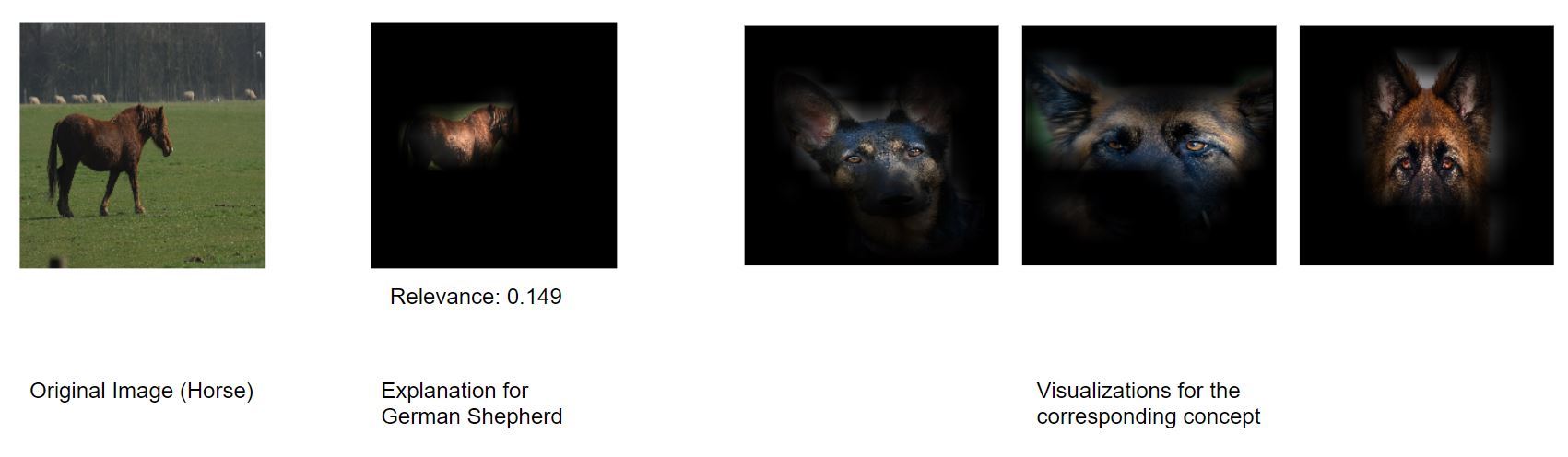}
    \caption{[Best viewed in color] The bulges on back and front side of horse make the region highlighted in the explanation for german shepherd look like ears (based on color). Since the concept itself is also of ear and eye region, this concept contributes towards german shepherd probability in this image.}
    \label{fig:horse_dog_color}
\end{figure}

\begin{figure}[H]
    \centering
    \includegraphics[width=\linewidth]{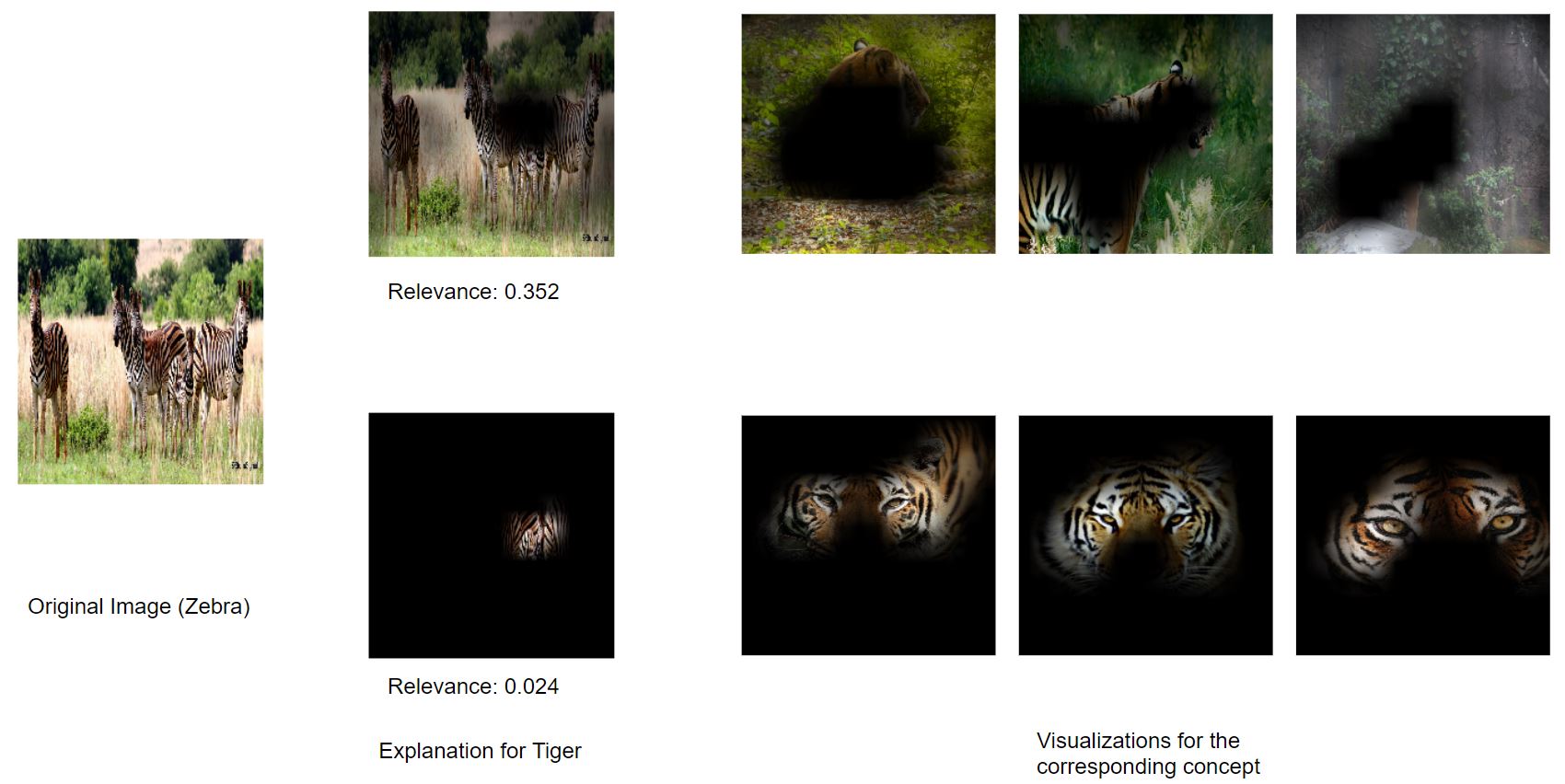}
    \caption{[Best viewed in color] In the concept in the first row, the model is looking at the background. Background of Zebra image and that of the images for the actual concept is quite similar. In the concept in second row, the model is looking at the stripes region in the zebra image, and the concept is also the stripes on tiger.}
    \label{fig:zebra_tiger}
\end{figure}

\begin{figure}[H]
    \centering
    \includegraphics[width=\linewidth]{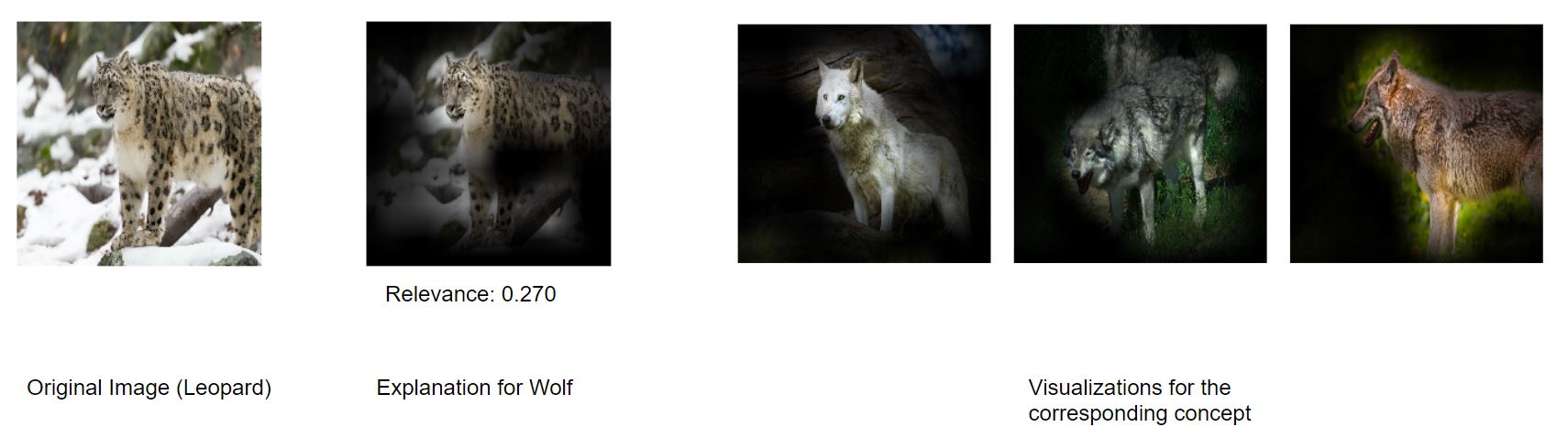}
    \caption{[Best viewed in color] The model is looking at the body of leopard in the explanation for wolf class. The concept itself also represents the body of wolf.}
    \label{fig:leopard_wolf_body}
\end{figure}

\begin{figure}[H]
    \centering
    \includegraphics[width=\linewidth]{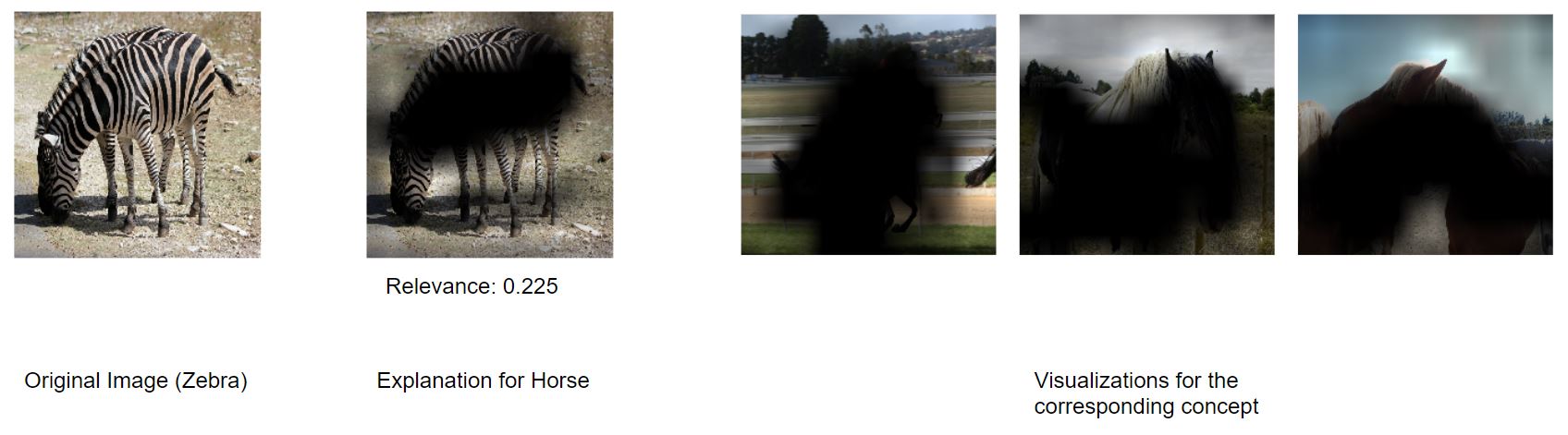}
    \caption{[Best viewed in color] The model is looking at the background. Background of Zebra image and that of the images for the actual concept is quite similar}
    \label{fig:zebra_background}
\end{figure}

\begin{figure}[H]
    \centering
    \includegraphics[width=\linewidth]{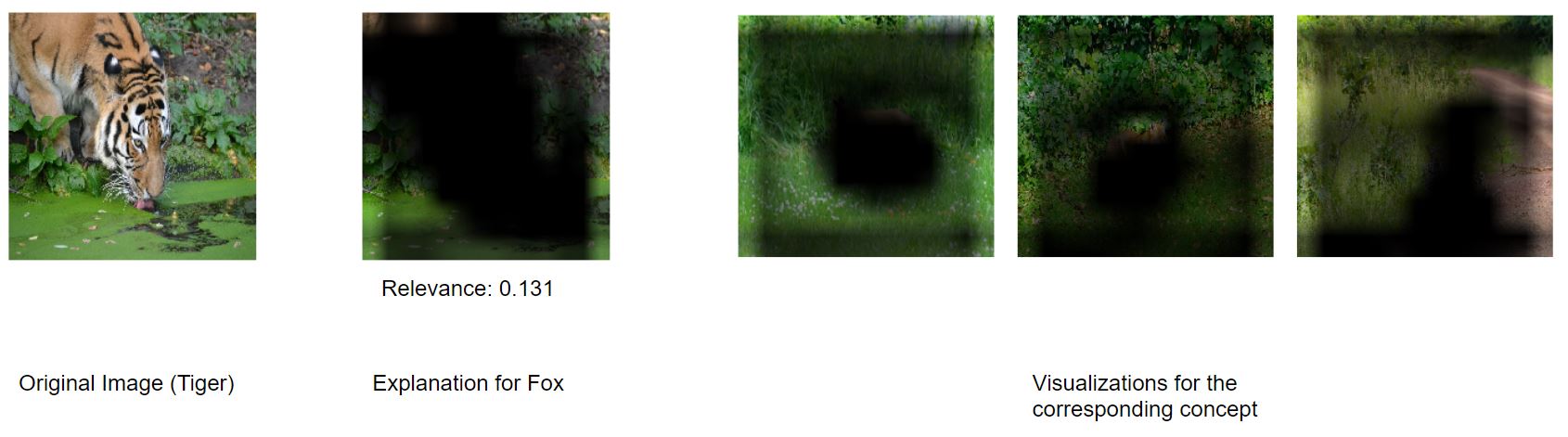}
    \caption{[Best viewed in color] The model is looking at the background. Background of Tiger image and that of the images for the actual concept is quite similar}
    \label{fig:tiger_background}
\end{figure}

\begin{figure}[H]
    \centering
    \includegraphics[width=\linewidth]{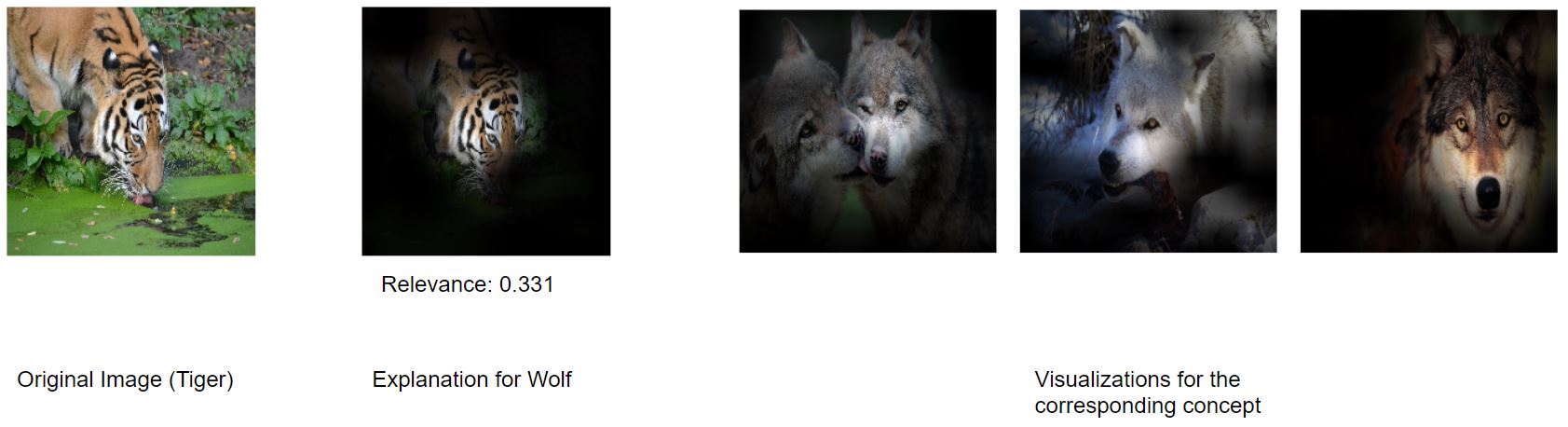}
    \caption{[Best viewed in color] The model is looking at the face of tiger in the explanation for wolf class. The concept itself also represents the face region of wolf}
    \label{fig:tiger_wolf_face}
\end{figure}

One of the major contributions of our method is that we can provide rich explanations for why a non-zero prediction probability is assigned to the class other than that of the predicted class. Figures ~\ref{fig:fox_lion_face}, ~\ref{fig:fox_dog_ear}, ~\ref{fig:horse_dog_color}, ~\ref{fig:zebra_tiger},  ~\ref{fig:leopard_wolf_body}, ~\ref{fig:zebra_background}, ~\ref{fig:tiger_background}, ~\ref{fig:tiger_wolf_face} show few such explanations.

\section{Stability}
\begin{figure}[H]
    \centering 
\begin{subfigure}{\textwidth}
  \includegraphics[width=0.12\linewidth]{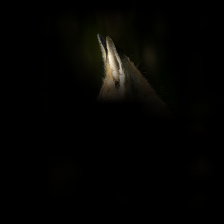}%
  \hfill
  \includegraphics[width=0.12\linewidth]{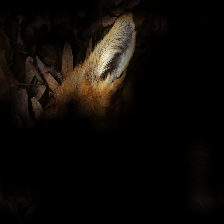}%
    \hfill
  \includegraphics[width=0.12\linewidth]{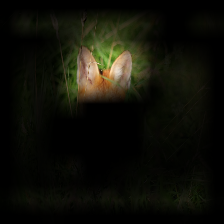}%
    \hfill
  \includegraphics[width=0.12\linewidth]{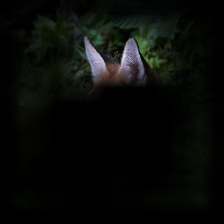}%
  \hfill
  \includegraphics[width=0.12\linewidth]{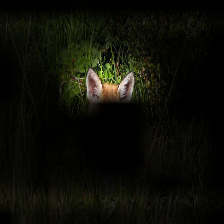}%
  \hfill
  \includegraphics[width=0.12\linewidth]{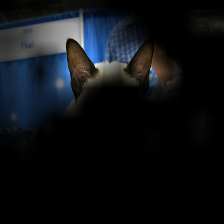}%
    \hfill
  \includegraphics[width=0.12\linewidth]{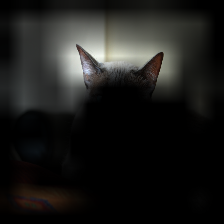}%
    \hfill
  \includegraphics[width=0.12\linewidth]{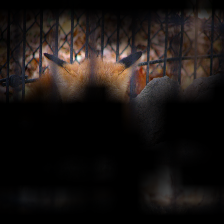}
  \caption{Fox - Ear}
  \label{fig:ear_fox}
\end{subfigure}
\\
\begin{subfigure}{\textwidth}
  \includegraphics[width=0.12\linewidth]{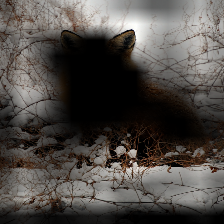}%
  \hfill
  \includegraphics[width=0.12\linewidth]{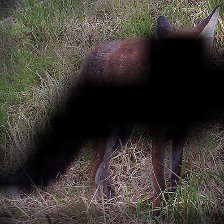}%
    \hfill
  \includegraphics[width=0.12\linewidth]{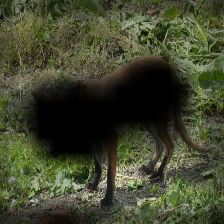}%
    \hfill
  \includegraphics[width=0.12\linewidth]{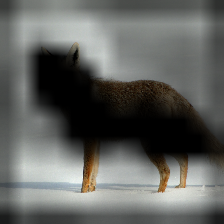}%
  \hfill
  \includegraphics[width=0.12\linewidth]{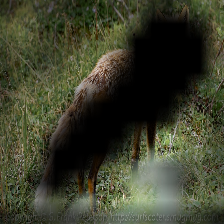}%
  \hfill
  \includegraphics[width=0.12\linewidth]{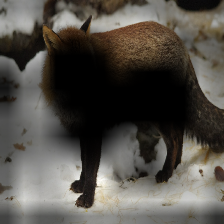}%
    \hfill
  \includegraphics[width=0.12\linewidth]{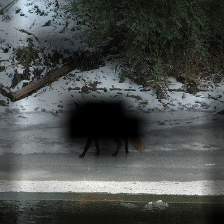}%
    \hfill
  \includegraphics[width=0.12\linewidth]{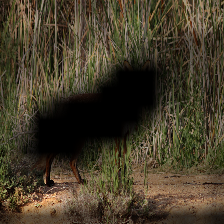}
  \caption{Fox - Background}
  \label{fig:background_fox}
\end{subfigure}
\\
\begin{subfigure}{\textwidth}
  \includegraphics[width=0.12\linewidth]{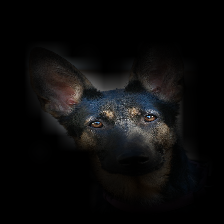}%
  \hfill
  \includegraphics[width=0.12\linewidth]{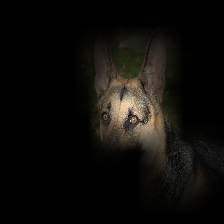}%
    \hfill
  \includegraphics[width=0.12\linewidth]{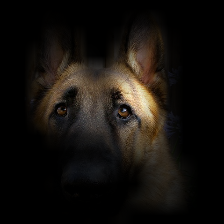}%
    \hfill
  \includegraphics[width=0.12\linewidth]{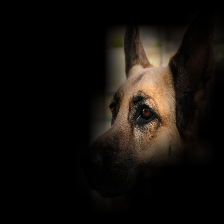}%
  \hfill
  \includegraphics[width=0.12\linewidth]{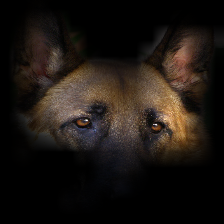}%
  \hfill
  \includegraphics[width=0.12\linewidth]{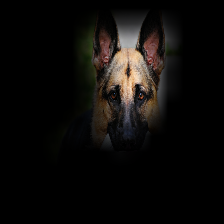}%
    \hfill
  \includegraphics[width=0.12\linewidth]{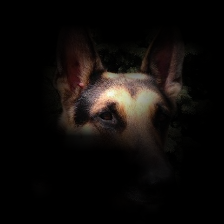}%
    \hfill
  \includegraphics[width=0.12\linewidth]{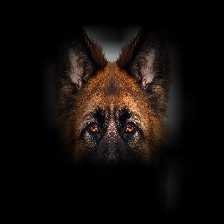}
  \caption{German Shepherd - Eyes}
  \label{fig:eyes_german_shepherd}
\end{subfigure}
\\
\begin{subfigure}{\textwidth}
  \includegraphics[width=0.12\linewidth]{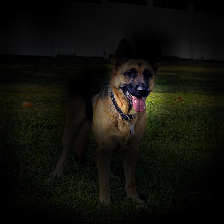}%
  \hfill
  \includegraphics[width=0.12\linewidth]{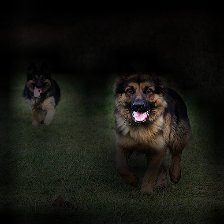}%
    \hfill
  \includegraphics[width=0.12\linewidth]{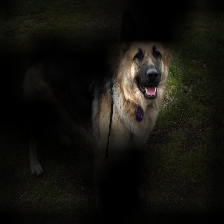}%
    \hfill
  \includegraphics[width=0.12\linewidth]{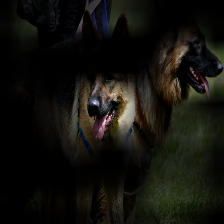}%
  \hfill
  \includegraphics[width=0.12\linewidth]{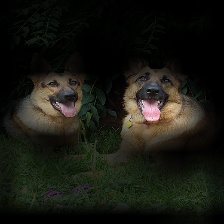}%
  \hfill
  \includegraphics[width=0.12\linewidth]{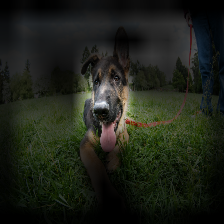}%
    \hfill
  \includegraphics[width=0.12\linewidth]{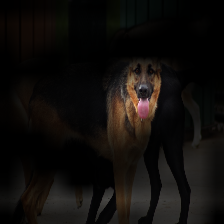}%
    \hfill
  \includegraphics[width=0.12\linewidth]{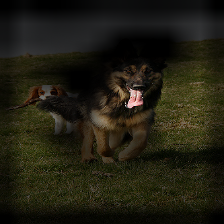}
  \caption{German Shepherd - Face}
  \label{fig:face_german_shepherd}
\end{subfigure}
\\
\begin{subfigure}{\textwidth}
  \includegraphics[width=0.12\linewidth]{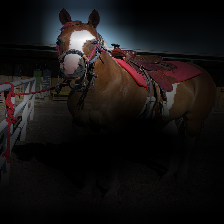}%
  \hfill
  \includegraphics[width=0.12\linewidth]{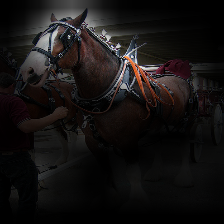}%
    \hfill
  \includegraphics[width=0.12\linewidth]{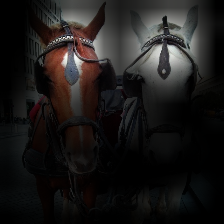}%
    \hfill
  \includegraphics[width=0.12\linewidth]{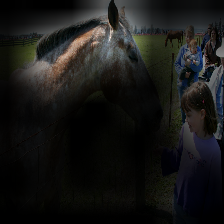}%
  \hfill
  \includegraphics[width=0.12\linewidth]{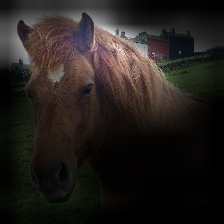}%
  \hfill
  \includegraphics[width=0.12\linewidth]{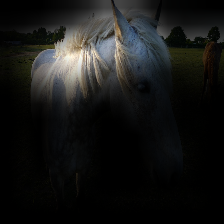}%
    \hfill
  \includegraphics[width=0.12\linewidth]{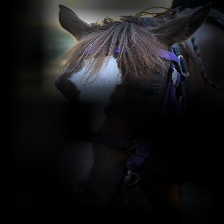}%
    \hfill
  \includegraphics[width=0.12\linewidth]{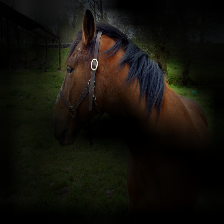}
  \caption{Horse - Crest}
  \label{fig:crest_horse}
\end{subfigure}
\\
\begin{subfigure}{\textwidth}
  \includegraphics[width=0.12\linewidth]{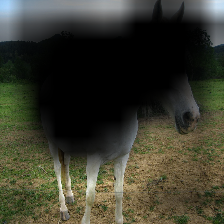}%
  \hfill
  \includegraphics[width=0.12\linewidth]{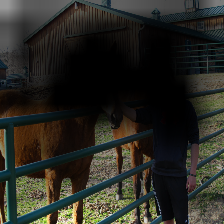}%
    \hfill
  \includegraphics[width=0.12\linewidth]{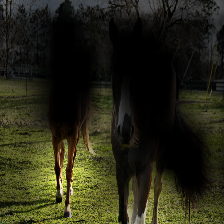}%
    \hfill
  \includegraphics[width=0.12\linewidth]{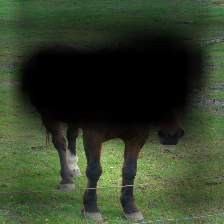}%
  \hfill
  \includegraphics[width=0.12\linewidth]{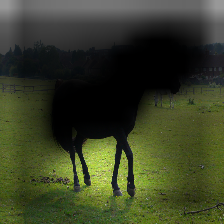}%
  \hfill
  \includegraphics[width=0.12\linewidth]{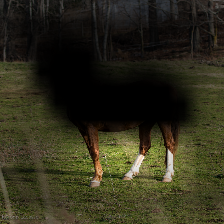}%
    \hfill
  \includegraphics[width=0.12\linewidth]{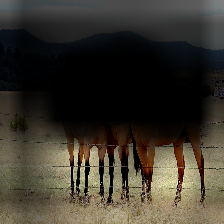}%
    \hfill
  \includegraphics[width=0.12\linewidth]{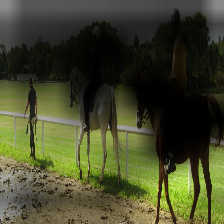}
  \caption{Horse - Legs}
  \label{fig:legs_horse}
\end{subfigure}
\\
\begin{subfigure}{\textwidth}
  \includegraphics[width=0.12\linewidth]{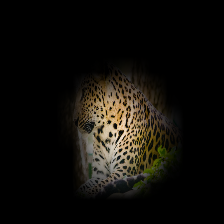}%
  \hfill
  \includegraphics[width=0.12\linewidth]{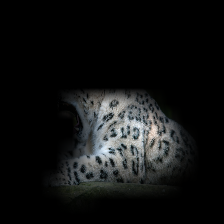}%
    \hfill
  \includegraphics[width=0.12\linewidth]{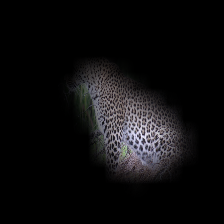}%
    \hfill
  \includegraphics[width=0.12\linewidth]{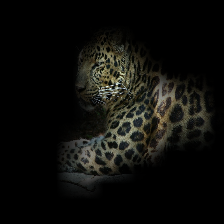}%
  \hfill
  \includegraphics[width=0.12\linewidth]{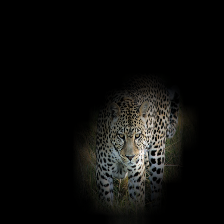}%
  \hfill
  \includegraphics[width=0.12\linewidth]{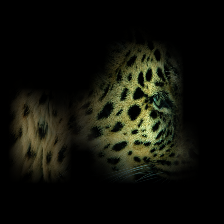}%
    \hfill
  \includegraphics[width=0.12\linewidth]{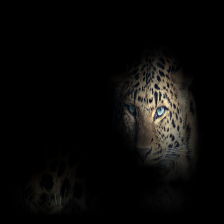}%
    \hfill
  \includegraphics[width=0.12\linewidth]{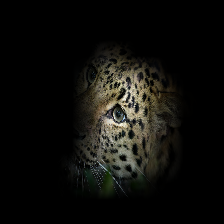}
  \caption{Leopard - Texture}
  \label{fig:texture_leopard}
\end{subfigure}
\\
\begin{subfigure}{\textwidth}
  \includegraphics[width=0.12\linewidth]{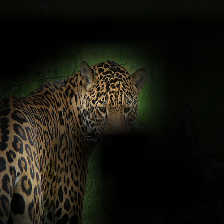}%
  \hfill
  \includegraphics[width=0.12\linewidth]{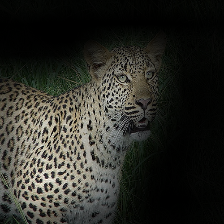}%
    \hfill
  \includegraphics[width=0.12\linewidth]{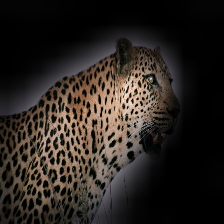}%
    \hfill
  \includegraphics[width=0.12\linewidth]{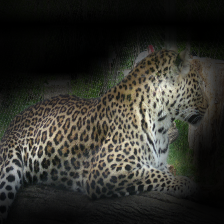}%
  \hfill
  \includegraphics[width=0.12\linewidth]{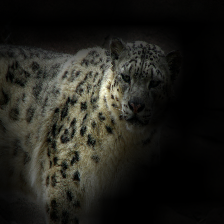}%
  \hfill
  \includegraphics[width=0.12\linewidth]{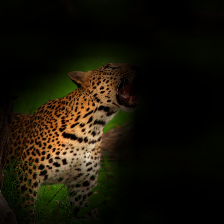}%
    \hfill
  \includegraphics[width=0.12\linewidth]{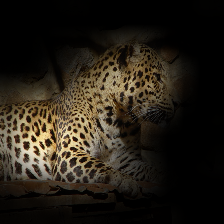}%
    \hfill
  \includegraphics[width=0.12\linewidth]{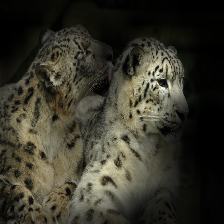}
  \caption{Leopard - Body}
  \label{fig:body_leopard}
\end{subfigure}
\\
\begin{subfigure}{\textwidth}
  \includegraphics[width=0.12\linewidth]{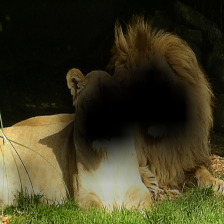}%
  \hfill
  \includegraphics[width=0.12\linewidth]{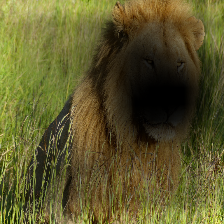}%
    \hfill
  \includegraphics[width=0.12\linewidth]{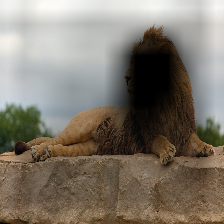}%
    \hfill
  \includegraphics[width=0.12\linewidth]{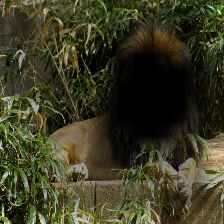}%
  \hfill
  \includegraphics[width=0.12\linewidth]{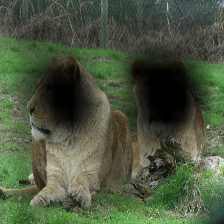}%
  \hfill
  \includegraphics[width=0.12\linewidth]{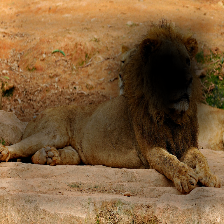}%
    \hfill
  \includegraphics[width=0.12\linewidth]{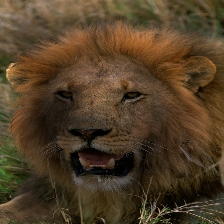}%
    \hfill
  \includegraphics[width=0.12\linewidth]{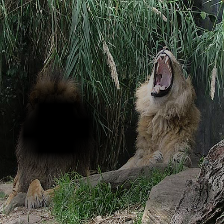}
  \caption{Lion - Body and Background}
  \label{fig:body_background_lion}
\end{subfigure}
\\
\begin{subfigure}{\textwidth}
  \includegraphics[width=0.12\linewidth]{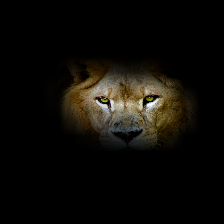}%
  \hfill
  \includegraphics[width=0.12\linewidth]{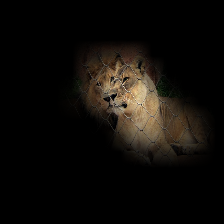}%
    \hfill
  \includegraphics[width=0.12\linewidth]{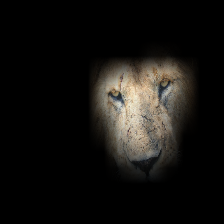}%
    \hfill
  \includegraphics[width=0.12\linewidth]{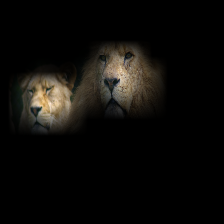}%
  \hfill
  \includegraphics[width=0.12\linewidth]{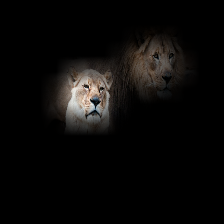}%
  \hfill
  \includegraphics[width=0.12\linewidth]{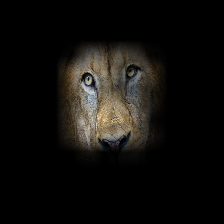}%
    \hfill
  \includegraphics[width=0.12\linewidth]{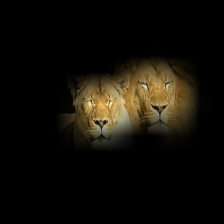}%
    \hfill
  \includegraphics[width=0.12\linewidth]{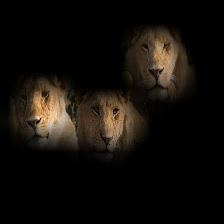}
  \caption{Lion - Face}
  \label{fig:face_lion}
\end{subfigure}
\caption{[Best viewed in color] Visualization of concepts across multiple images of the same class}
\label{fig:outputIllustration1}
\end{figure}

\begin{figure}[H]
    \centering 
\begin{subfigure}{\textwidth}
  \includegraphics[width=0.12\linewidth]{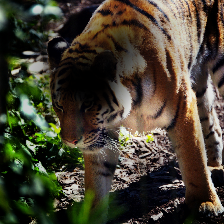}%
  \hfill
  \includegraphics[width=0.12\linewidth]{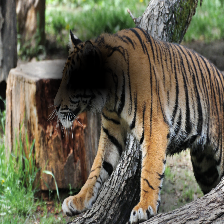}%
    \hfill
  \includegraphics[width=0.12\linewidth]{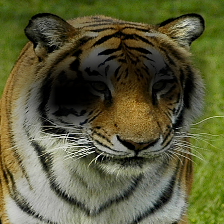}%
    \hfill
  \includegraphics[width=0.12\linewidth]{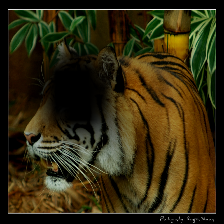}%
  \hfill
  \includegraphics[width=0.12\linewidth]{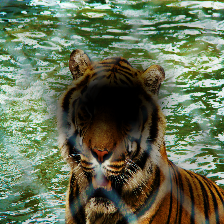}%
  \hfill
  \includegraphics[width=0.12\linewidth]{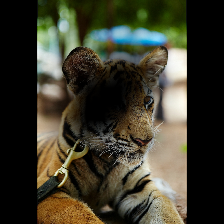}%
    \hfill
  \includegraphics[width=0.12\linewidth]{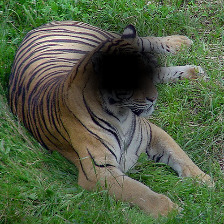}%
    \hfill
  \includegraphics[width=0.12\linewidth]{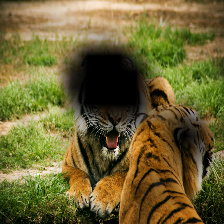}
  \caption{Tiger - Body and Background }
  \label{fig:body_background_tiger}
\end{subfigure}
\\
\begin{subfigure}{\textwidth}
  \includegraphics[width=0.12\linewidth]{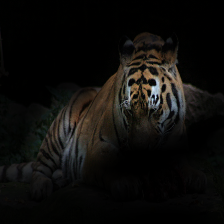}%
  \hfill
  \includegraphics[width=0.12\linewidth]{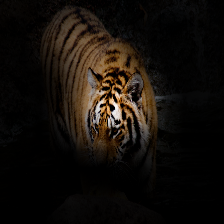}%
    \hfill
  \includegraphics[width=0.12\linewidth]{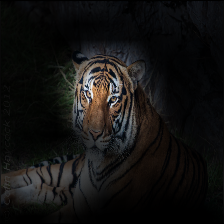}%
    \hfill
  \includegraphics[width=0.12\linewidth]{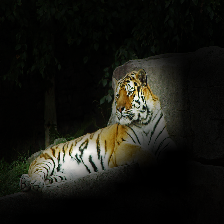}%
  \hfill
  \includegraphics[width=0.12\linewidth]{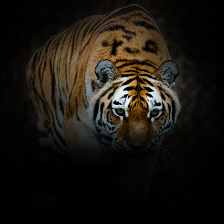}%
  \hfill
  \includegraphics[width=0.12\linewidth]{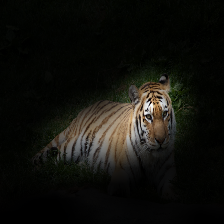}%
    \hfill
  \includegraphics[width=0.12\linewidth]{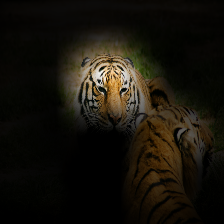}%
    \hfill
  \includegraphics[width=0.12\linewidth]{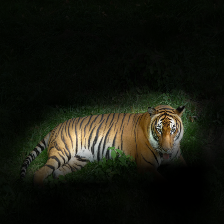}
  \caption{Tiger - Face \& Upper Body}
  \label{fig:face_upper_body_tiger}
\end{subfigure}
\\
\begin{subfigure}{\textwidth}
  \includegraphics[width=0.12\linewidth]{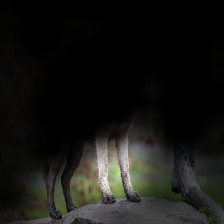}%
  \hfill
  \includegraphics[width=0.12\linewidth]{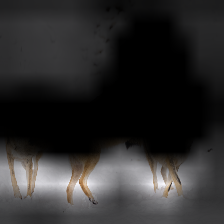}%
    \hfill
  \includegraphics[width=0.12\linewidth]{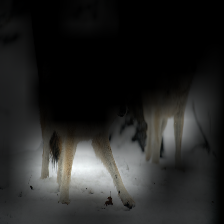}%
    \hfill
  \includegraphics[width=0.12\linewidth]{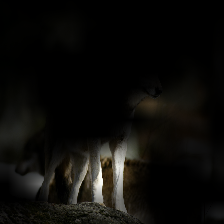}%
  \hfill
  \includegraphics[width=0.12\linewidth]{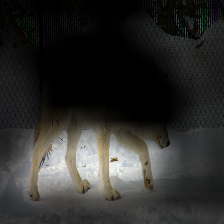}%
  \hfill
  \includegraphics[width=0.12\linewidth]{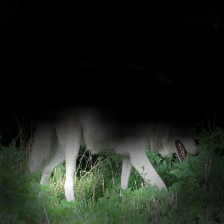}%
    \hfill
  \includegraphics[width=0.12\linewidth]{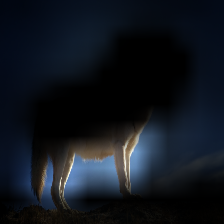}%
    \hfill
  \includegraphics[width=0.12\linewidth]{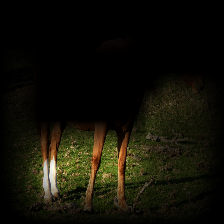}
  \caption{Wolf - Legs }
  \label{fig:legs_wolf}
\end{subfigure}
\\
\begin{subfigure}{\textwidth}
  \includegraphics[width=0.12\linewidth]{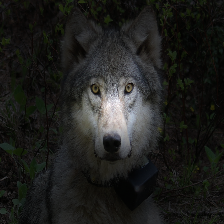}%
  \hfill
  \includegraphics[width=0.12\linewidth]{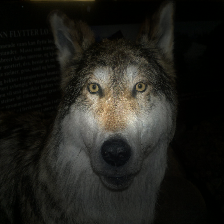}%
    \hfill
  \includegraphics[width=0.12\linewidth]{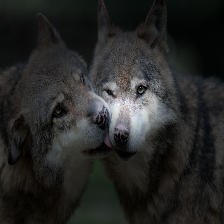}%
    \hfill
  \includegraphics[width=0.12\linewidth]{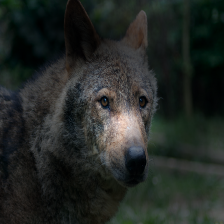}%
  \hfill
  \includegraphics[width=0.12\linewidth]{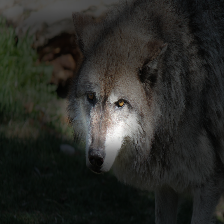}%
  \hfill
  \includegraphics[width=0.12\linewidth]{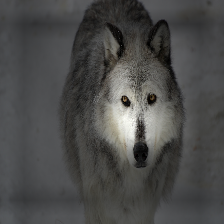}%
    \hfill
  \includegraphics[width=0.12\linewidth]{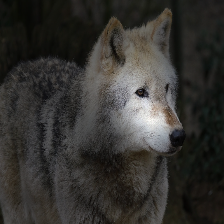}%
    \hfill
  \includegraphics[width=0.12\linewidth]{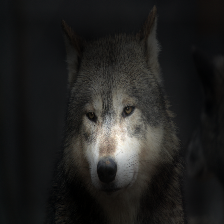}
  \caption{Wolf - Muzzle}
  \label{fig:muzzle_wolf}
\end{subfigure}
\\
\begin{subfigure}{\textwidth}
  \includegraphics[width=0.12\linewidth]{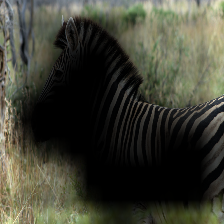}%
  \hfill
  \includegraphics[width=0.12\linewidth]{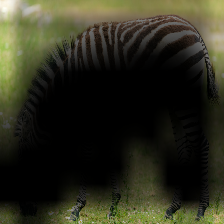}%
    \hfill
  \includegraphics[width=0.12\linewidth]{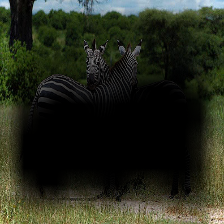}%
    \hfill
  \includegraphics[width=0.12\linewidth]{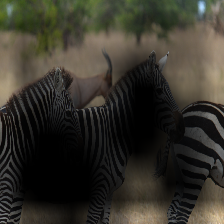}%
  \hfill
  \includegraphics[width=0.12\linewidth]{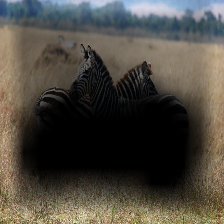}%
  \hfill
  \includegraphics[width=0.12\linewidth]{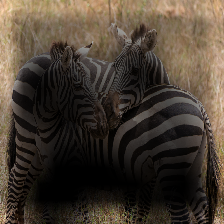}%
    \hfill
  \includegraphics[width=0.12\linewidth]{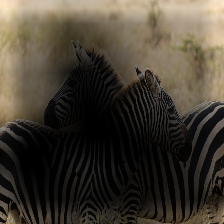}%
    \hfill
  \includegraphics[width=0.12\linewidth]{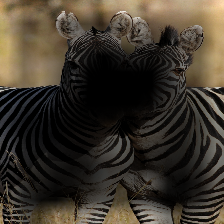}
  \caption{Zebra - Background}
  \label{fig:background_zebra}
\end{subfigure}
\\
\begin{subfigure}{\textwidth}
  \includegraphics[width=0.12\linewidth]{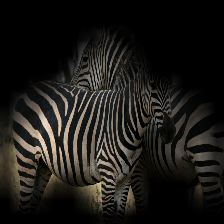}%
  \hfill
  \includegraphics[width=0.12\linewidth]{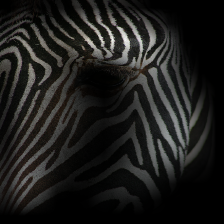}%
    \hfill
  \includegraphics[width=0.12\linewidth]{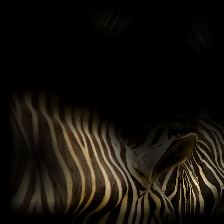}%
    \hfill
  \includegraphics[width=0.12\linewidth]{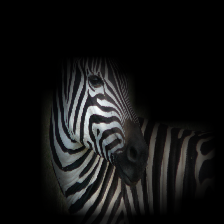}%
  \hfill
  \includegraphics[width=0.12\linewidth]{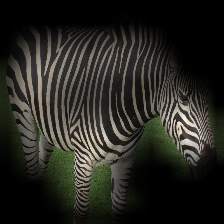}%
  \hfill
  \includegraphics[width=0.12\linewidth]{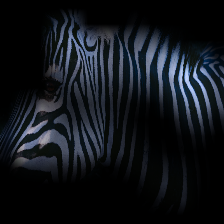}%
    \hfill
  \includegraphics[width=0.12\linewidth]{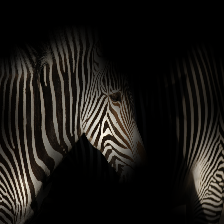}%
    \hfill
  \includegraphics[width=0.12\linewidth]{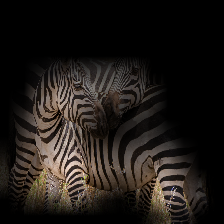}
  \caption{Zebra - Stripes}
  \label{fig:stripes_zebra}
\end{subfigure}
\caption{[Best viewed in color] Visualization of concepts across multiple images of the same class}
\label{fig:outputIllustration2}
\end{figure}

\begin{figure}[H]
    \centering
    \includegraphics[width=0.45\textwidth]{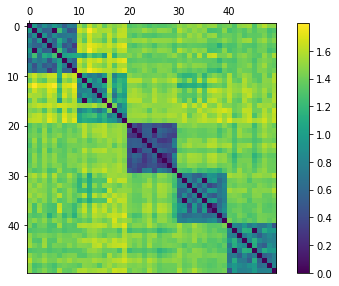}
    \caption{[Best viewed in color] Pairwise Euclidean distance between concept embeddings}
    \label{fig:eucl_dis}
\end{figure}

Stability of an interpretation refers to visual comparison of the learned concepts for images of the same class with varying orientation, size or position. The explanations of such images should be similar as well i.e., the learned concepts should be consistent. Given the saliency maps of a few such images, we should be able to understand the underlying concept easily. Figures \ref{fig:outputIllustration1} and \ref{fig:outputIllustration2} show that our learned concepts have very high stability. We also demonstrate that the concept embeddings for a particular concept for images with varying orientation, size or position are very close to each other. We take a small set of 10 images from the fox class and randomly choose 5 concepts of this class. For each concept we calculate the pairwise Euclidean distance between the concept embedding of the images. The results are shown in Figure \ref{fig:eucl_dis} in the form of a box diagonal matrix. Let $c_i$ denote the concept and $f_i$ denote the image then the axis of the matrix are defined as $[(c_1,f_1),..(c_1,f_{10})), (c_2,f_1),..(c_2,f_{10}),..(c_5,f_1)..(c_5,f_{10})]$. As it can be seen that the distances along the diagonal is lesser compared to the non-diagonal distances. This shows that the embeddings of similar (dissimilar) concepts are closer (farther).

\section{Comparison of robustness}

\begin{figure}[H]
    \centering 
\begin{subfigure}{\textwidth}
    \includegraphics[width=0.24\linewidth]{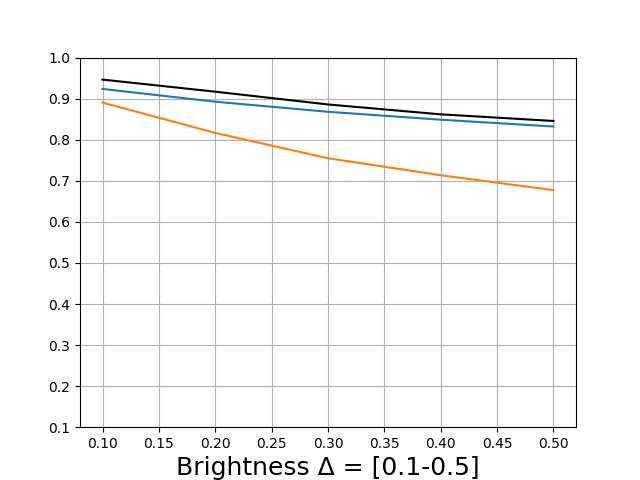}%
    \includegraphics[width=0.24\linewidth]{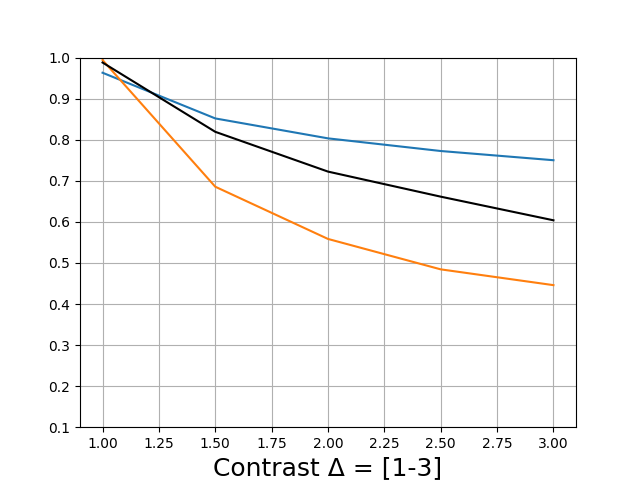}%
  \includegraphics[width=0.24\linewidth]{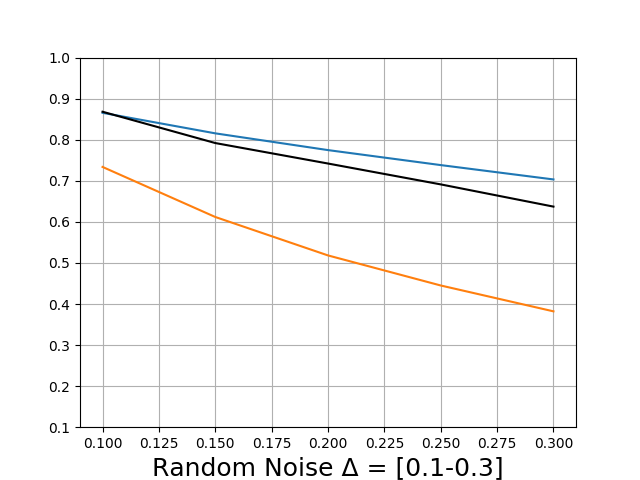}%
  \includegraphics[width=0.24\linewidth]{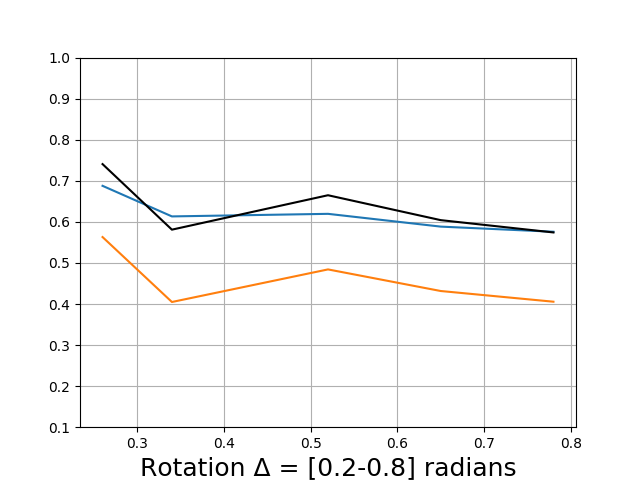}
\caption{Threshold = 0.4}
\end{subfigure}
\\
\begin{subfigure}{\textwidth}
  \includegraphics[width=0.24\linewidth]{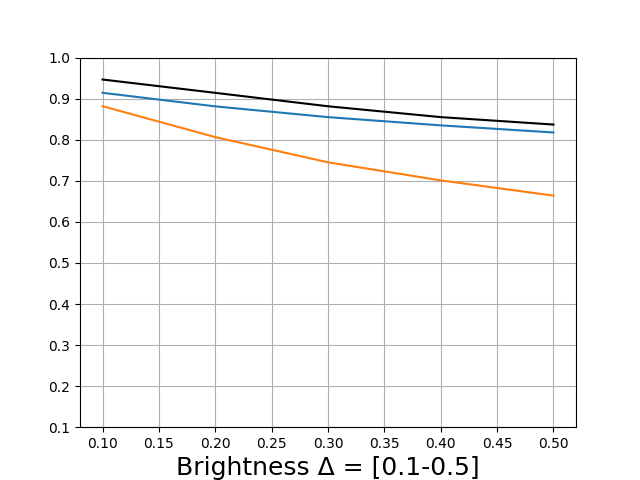}%
  \includegraphics[width=0.24\linewidth]{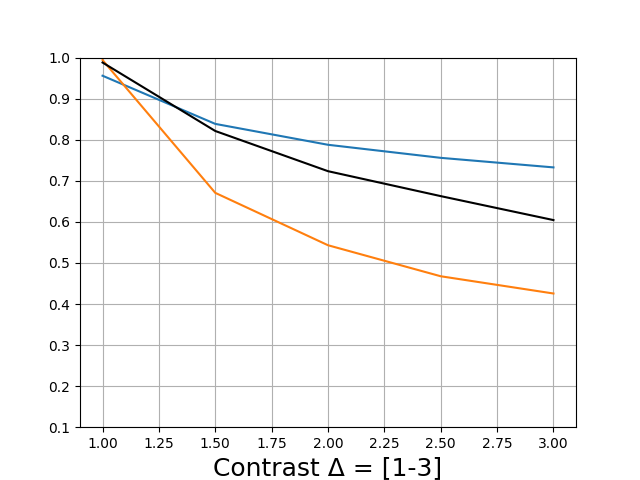}%
  \includegraphics[width=0.24\linewidth]{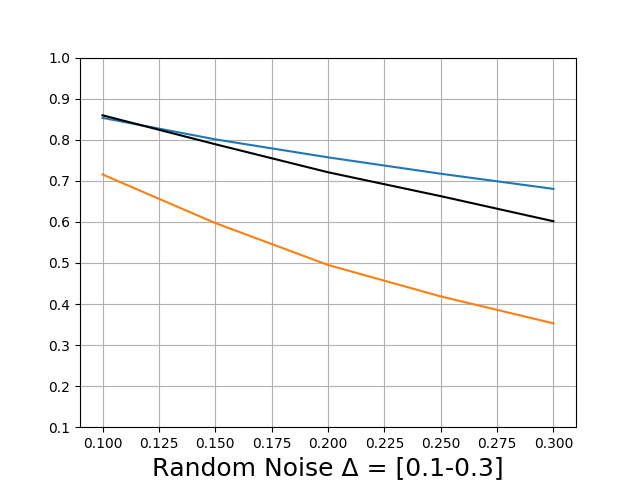}%
  \includegraphics[width=0.24\linewidth]{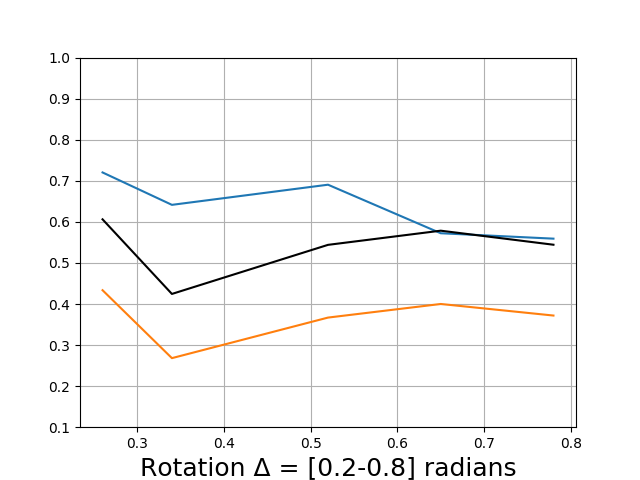}
  \caption{Threshold = 0.5}
\end{subfigure}
\\
\begin{subfigure}{\textwidth}
  \includegraphics[width=0.24\linewidth]{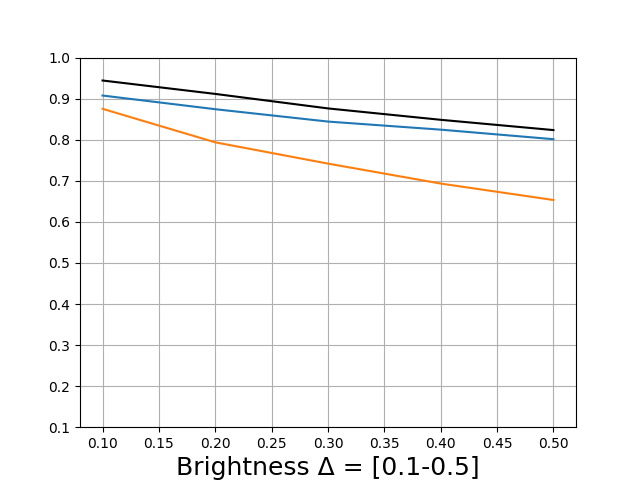}%
  \includegraphics[width=0.24\linewidth]{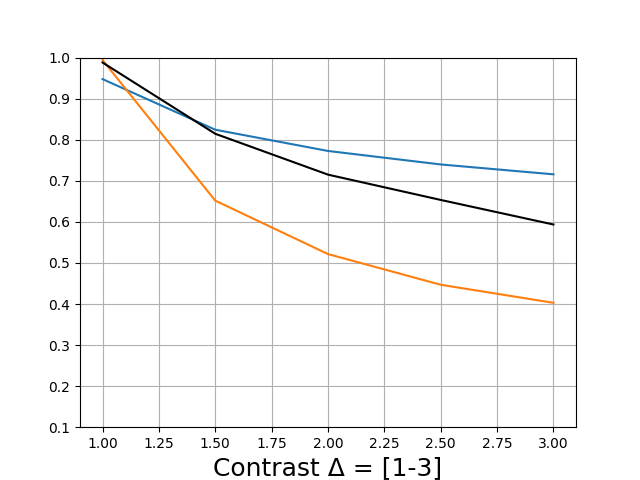}%
  \includegraphics[width=0.24\linewidth]{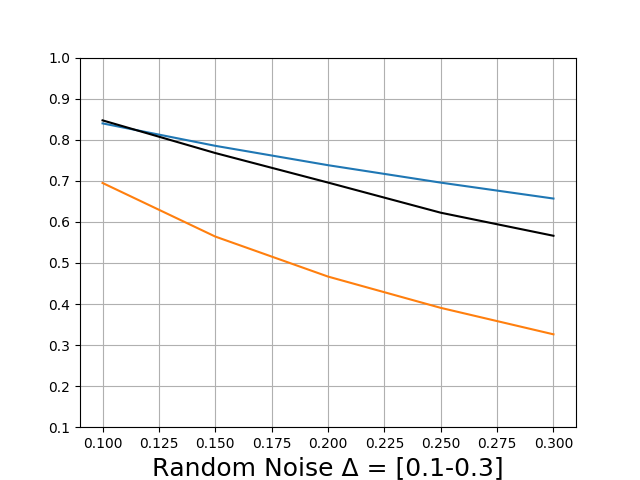}%
  \includegraphics[width=0.24\linewidth]{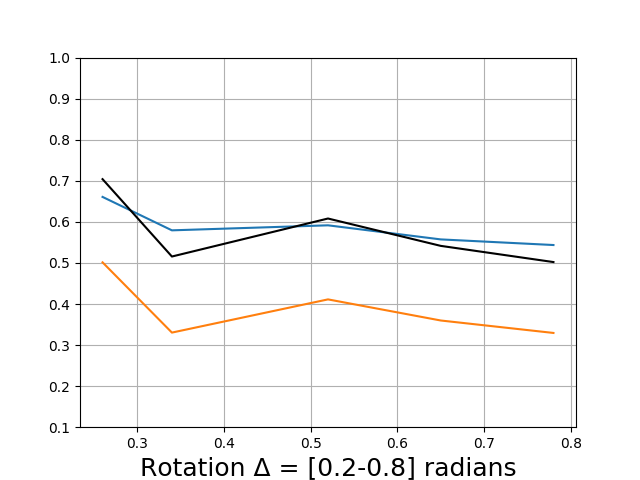}
  \caption{Threshold = 0.6}
\end{subfigure}

\includegraphics[width=0.5\textwidth]{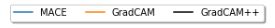}
\caption{[Best viewed in color] Intersection Over Union for thresholded saliency maps of original and perturbed image }
\label{fig:robustness_exp}
\end{figure}

Robustness is an important desideratum of an explanation. The explanation generated by any interpretability method should be robust to local perturbations in the input image. Figure \ref{fig:robustness_exp} shows that this is not the case for popular interpretability methods; even adding minimal noise to the input introduces visible changes in the explanations. We formally quantify this parameter of evaluation as the intersection over union between the explanations of the image and its perturbed variation:
\begin{equation}
R(x_{i}) = IoU( f_{expl}(x_{i}) , f_{expl}(\tilde{x}_{i}) )
\end{equation}
Here $x_i$ is an input image and $\tilde{x}_i$ is the perturbed image. $f_{expl}$ refers to the explanation function whose output is the saliency map for the interpretation method. The saliency maps are thresholded to create a binary map before calculating the IoU. We perform robustness experiment on various parameters including variations in brightness, contrast, random noise and rotation. For each parameter we define a range of values that describes the intensity of perturbation, eg. standard deviation of the noise distribution to be added in an image, delta value to increase the brightness/contrast in the image, angle of rotation etc. We define a wide range of threshold values in the range $[0.3-0.7]$ to generate the binary maps and compare their robustness for three approaches including MACE, GradCAM \cite{gradCAM} and GradCAM++ \cite{grad_cam_plus_plus}. 

\section{Places Experiments}
\begin{figure}[H]
    \centering 
\begin{subfigure}{\textwidth}
  \includegraphics[width=0.12\linewidth]{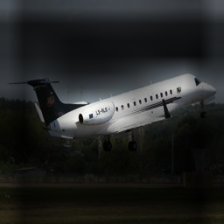}%
  \hfill
  \includegraphics[width=0.12\linewidth]{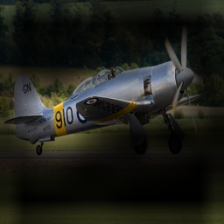}%
    \hfill
  \includegraphics[width=0.12\linewidth]{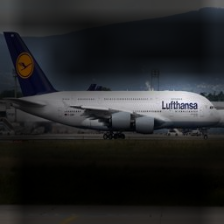}%
    \hfill
  \includegraphics[width=0.12\linewidth]{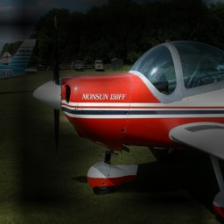}%
  \hfill
  \includegraphics[width=0.12\linewidth]{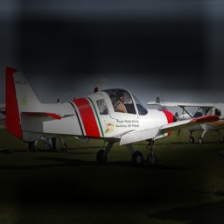}%
  \hfill
  \includegraphics[width=0.12\linewidth]{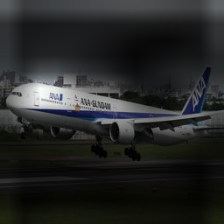}%
    \hfill
  \includegraphics[width=0.12\linewidth]{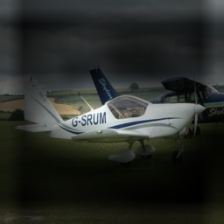}%
    \hfill
  \includegraphics[width=0.12\linewidth]{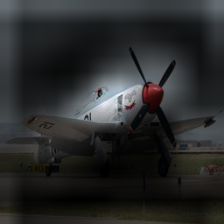}
  \label{fig:airfield_airplane}
  \caption{Airfield - Airplane concept}
\end{subfigure}
\begin{subfigure}{\textwidth}
  \includegraphics[width=0.12\linewidth]{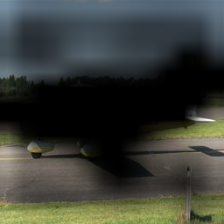}%
  \hfill
  \includegraphics[width=0.12\linewidth]{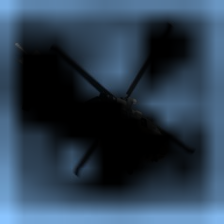}%
    \hfill
  \includegraphics[width=0.12\linewidth]{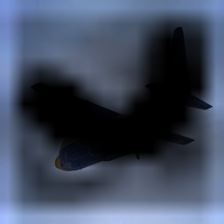}%
    \hfill
  \includegraphics[width=0.12\linewidth]{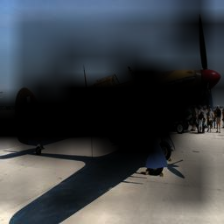}%
  \hfill
  \includegraphics[width=0.12\linewidth]{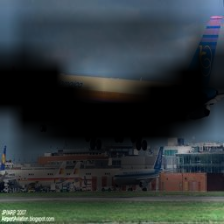}%
  \hfill
  \includegraphics[width=0.12\linewidth]{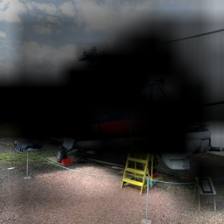}%
    \hfill
  \includegraphics[width=0.12\linewidth]{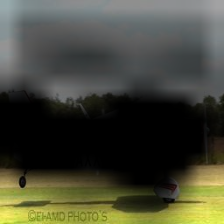}%
    \hfill
  \includegraphics[width=0.12\linewidth]{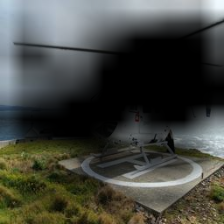}
  \label{fig:airfield_background}
  \caption{Airfield - Background}
\end{subfigure}
\begin{subfigure}{\textwidth}
  \includegraphics[width=0.12\linewidth]{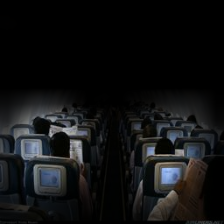}%
  \hfill
  \includegraphics[width=0.12\linewidth]{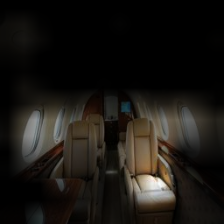}%
    \hfill
  \includegraphics[width=0.12\linewidth]{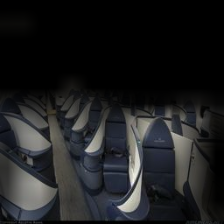}%
    \hfill
  \includegraphics[width=0.12\linewidth]{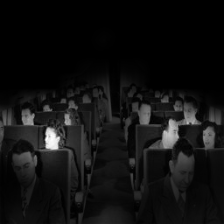}%
  \hfill
  \includegraphics[width=0.12\linewidth]{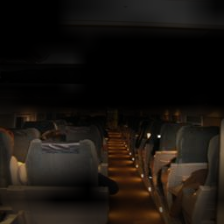}%
  \hfill
  \includegraphics[width=0.12\linewidth]{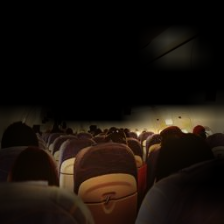}%
    \hfill
  \includegraphics[width=0.12\linewidth]{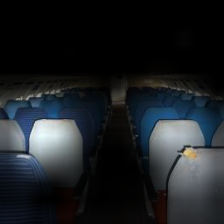}%
    \hfill
  \includegraphics[width=0.12\linewidth]{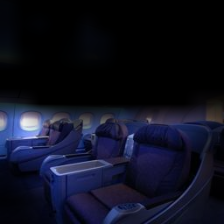}
  \label{fig:airplane_cabin_seats}
  \caption{Airplane Cabin - seats in the airplane cabin}
\end{subfigure}
\begin{subfigure}{\textwidth}
  \includegraphics[width=0.12\linewidth]{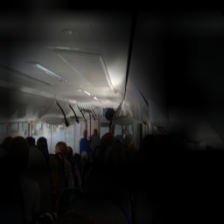}%
  \hfill
  \includegraphics[width=0.12\linewidth]{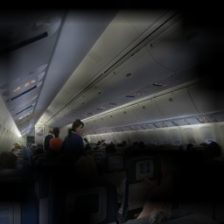}%
    \hfill
  \includegraphics[width=0.12\linewidth]{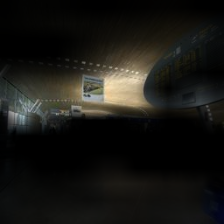}%
    \hfill
  \includegraphics[width=0.12\linewidth]{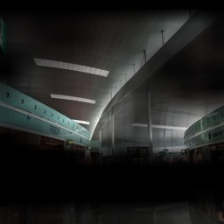}%
  \hfill
  \includegraphics[width=0.12\linewidth]{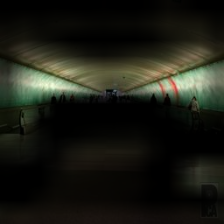}%
  \hfill
  \includegraphics[width=0.12\linewidth]{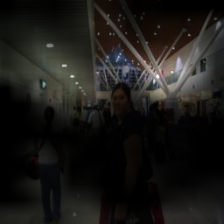}%
    \hfill
  \includegraphics[width=0.12\linewidth]{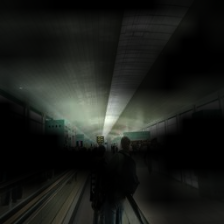}%
    \hfill
  \includegraphics[width=0.12\linewidth]{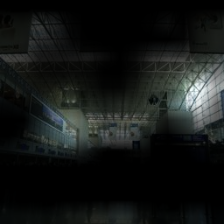}
  \label{fig:airplane_cabin_top_lights}
  \caption{Airplane Cabin - Lights on top in airplane cabin}
\end{subfigure}
\begin{subfigure}{\textwidth}
  \includegraphics[width=0.12\linewidth]{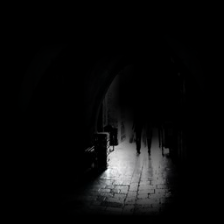}%
  \hfill
  \includegraphics[width=0.12\linewidth]{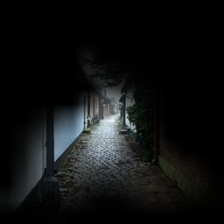}%
    \hfill
  \includegraphics[width=0.12\linewidth]{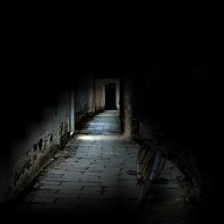}%
    \hfill
  \includegraphics[width=0.12\linewidth]{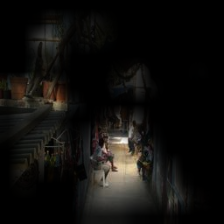}%
  \hfill
  \includegraphics[width=0.12\linewidth]{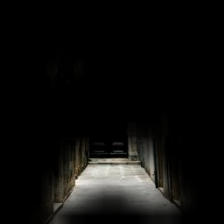}%
  \hfill
  \includegraphics[width=0.12\linewidth]{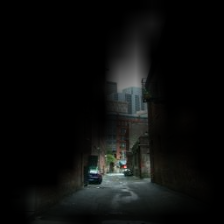}%
    \hfill
  \includegraphics[width=0.12\linewidth]{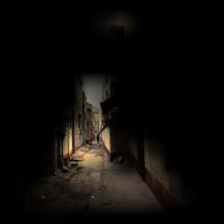}%
    \hfill
  \includegraphics[width=0.12\linewidth]{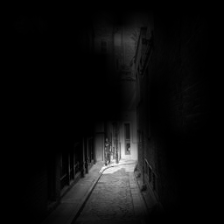}
  \label{fig:alley_light_region}
  \caption{Alley - Region with light coming from the end of alley}
\end{subfigure}
\begin{subfigure}{\textwidth}
  \includegraphics[width=0.12\linewidth]{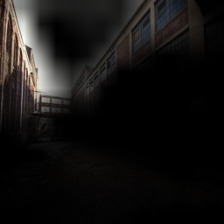}%
  \hfill
  \includegraphics[width=0.12\linewidth]{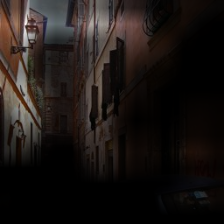}%
    \hfill
  \includegraphics[width=0.12\linewidth]{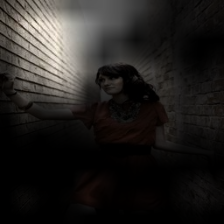}%
    \hfill
  \includegraphics[width=0.12\linewidth]{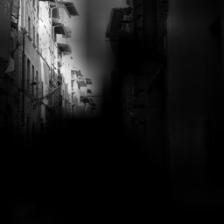}%
  \hfill
  \includegraphics[width=0.12\linewidth]{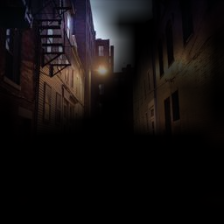}%
  \hfill
  \includegraphics[width=0.12\linewidth]{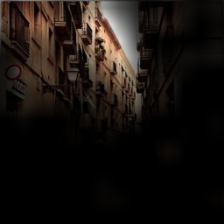}%
    \hfill
  \includegraphics[width=0.12\linewidth]{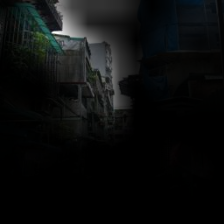}%
    \hfill
  \includegraphics[width=0.12\linewidth]{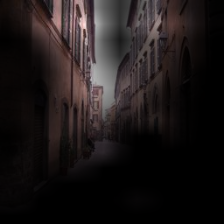}
  \label{fig:alley_walls}
  \caption{Alley - Walls of houses}
\end{subfigure}
\begin{subfigure}{\textwidth}
  \includegraphics[width=0.12\linewidth]{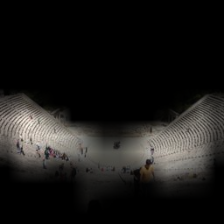}%
  \hfill
  \includegraphics[width=0.12\linewidth]{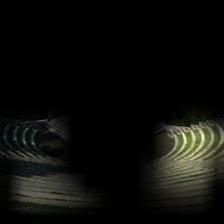}%
    \hfill
  \includegraphics[width=0.12\linewidth]{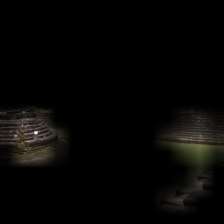}%
    \hfill
  \includegraphics[width=0.12\linewidth]{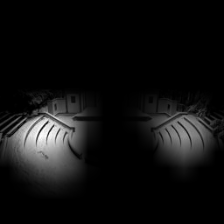}%
  \hfill
  \includegraphics[width=0.12\linewidth]{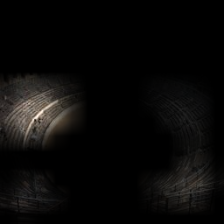}%
  \hfill
  \includegraphics[width=0.12\linewidth]{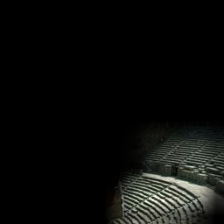}%
    \hfill
  \includegraphics[width=0.12\linewidth]{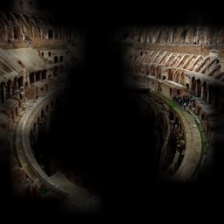}%
    \hfill
  \includegraphics[width=0.12\linewidth]{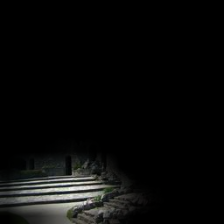}
  \label{fig:amphitheater_curvature_steps}
  \caption{Amphitheater - The curvature and steps in amphitheater}
\end{subfigure}
\begin{subfigure}{\textwidth}
  \includegraphics[width=0.12\linewidth]{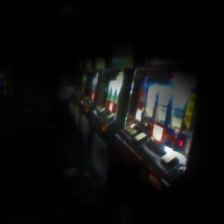}%
  \hfill
  \includegraphics[width=0.12\linewidth]{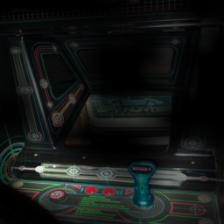}%
    \hfill
  \includegraphics[width=0.12\linewidth]{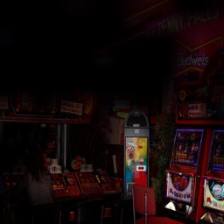}%
    \hfill
  \includegraphics[width=0.12\linewidth]{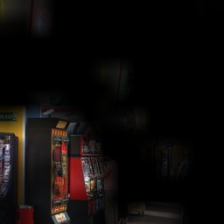}%
  \hfill
  \includegraphics[width=0.12\linewidth]{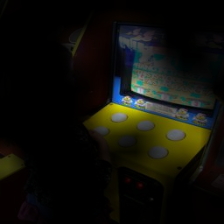}%
  \hfill
  \includegraphics[width=0.12\linewidth]{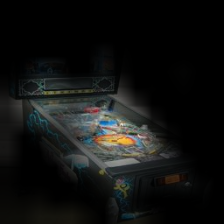}%
    \hfill
  \includegraphics[width=0.12\linewidth]{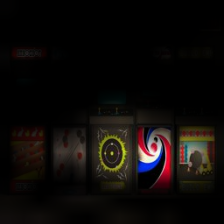}%
    \hfill
  \includegraphics[width=0.12\linewidth]{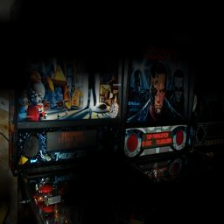}
  \label{fig:amusement_arcade_games}
  \caption{Amusement Arcade - Games region in amusement arcade}
\end{subfigure}
\caption{[Best viewed in color] Concepts for Places365 Dataset.}
\label{fig:places_365_concepts}
\end{figure}

The visualizations for the VGG model, trained on Places365 are shown in Figure \ref{fig:places_365_concepts}. We observed that most of the concepts were pruned and we had just 2-3 useful concepts per class. One of the reasons could be that the classes in the dataset were different from each other and didn't require a large set of concepts to make the prediction.

\section{ResNet Experiments}
\begin{figure}[H]
    \centering 
\begin{subfigure}{\textwidth}
  \includegraphics[width=0.12\linewidth]{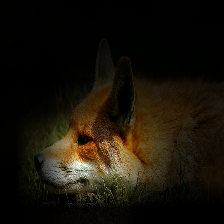}%
  \hfill
  \includegraphics[width=0.12\linewidth]{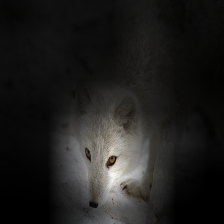}%
    \hfill
  \includegraphics[width=0.12\linewidth]{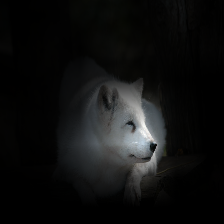}%
    \hfill
  \includegraphics[width=0.12\linewidth]{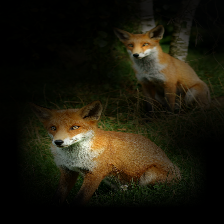}%
  \hfill
  \includegraphics[width=0.12\linewidth]{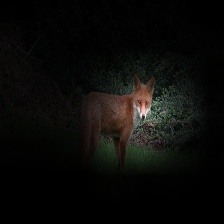}%
  \hfill
  \includegraphics[width=0.12\linewidth]{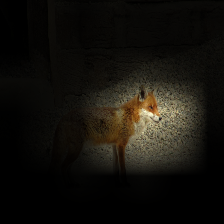}%
  \hfill
  \includegraphics[width=0.12\linewidth]{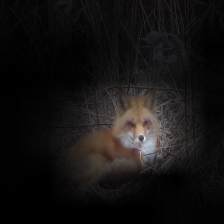}%
   \hfill
  \includegraphics[width=0.12\linewidth]{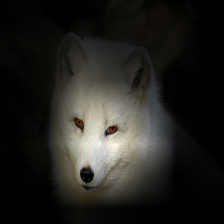}%
    
  \caption{Fox - Head}
  \label{fig:fox_head}
\end{subfigure}
\\
\begin{subfigure}{\textwidth}
  \includegraphics[width=0.12\linewidth]{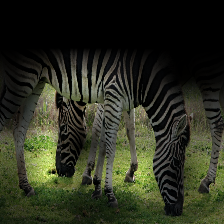}%
  \hfill
  \includegraphics[width=0.12\linewidth]{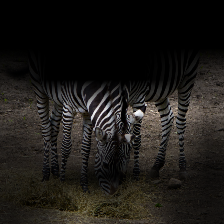}%
    \hfill
  \includegraphics[width=0.12\linewidth]{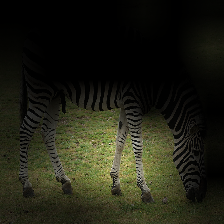}%
    \hfill
  \includegraphics[width=0.12\linewidth]{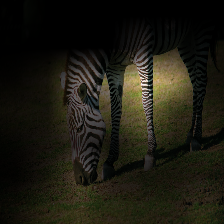}%
  \hfill
  \includegraphics[width=0.12\linewidth]{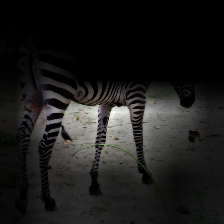}%
  \hfill
  \includegraphics[width=0.12\linewidth]{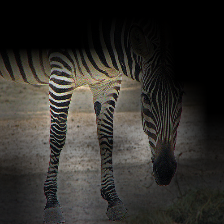}%
  \hfill
  \includegraphics[width=0.12\linewidth]{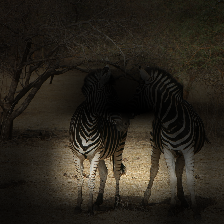}%
  \hfill
  \includegraphics[width=0.12\linewidth]{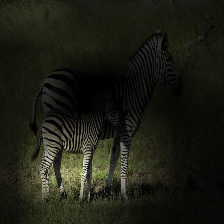}%
   
  \caption{Zebra-Legs}
  \label{fig:zebra_legs}
\end{subfigure}
\\
\begin{subfigure}{\textwidth}
  \includegraphics[width=0.12\linewidth]{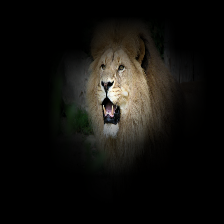}%
  \hfill
  \includegraphics[width=0.12\linewidth]{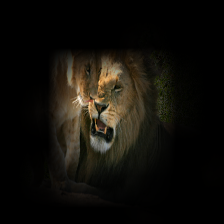}%
    \hfill
  \includegraphics[width=0.12\linewidth]{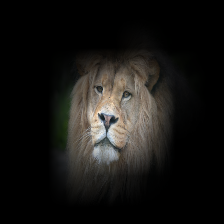}%
    \hfill
  \includegraphics[width=0.12\linewidth]{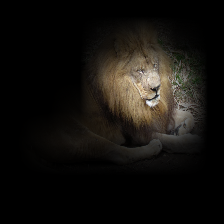}%
  \hfill
  \includegraphics[width=0.12\linewidth]{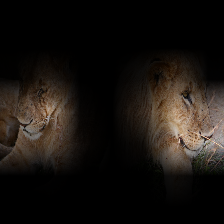}%
  \hfill
  \includegraphics[width=0.12\linewidth]{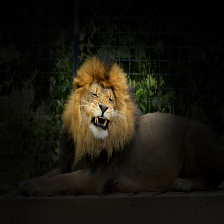}%
  \hfill
  \includegraphics[width=0.12\linewidth]{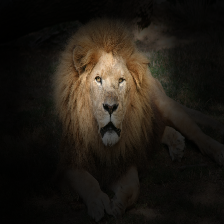}%
  \hfill
  \includegraphics[width=0.12\linewidth]{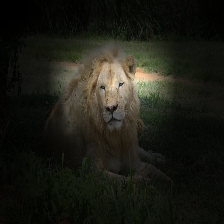}%
   
  \caption{Lion-Head}
  \label{fig:lion_head}
\end{subfigure}
\\
\begin{subfigure}{\textwidth}
  \includegraphics[width=0.12\linewidth]{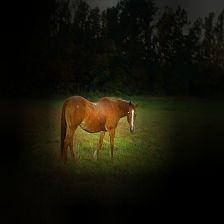}%
  \hfill
  \includegraphics[width=0.12\linewidth]{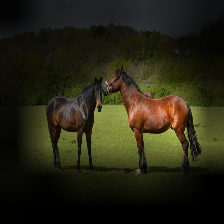}%
    \hfill
  \includegraphics[width=0.12\linewidth]{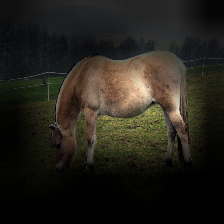}%
    \hfill
  \includegraphics[width=0.12\linewidth]{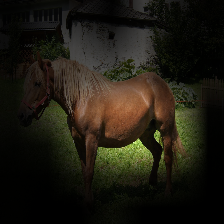}%
  \hfill
  \includegraphics[width=0.12\linewidth]{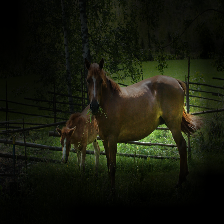}%
  \hfill
  \includegraphics[width=0.12\linewidth]{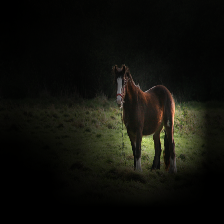}%
  \hfill
  \includegraphics[width=0.12\linewidth]{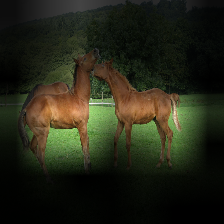}%
  \hfill
  \includegraphics[width=0.12\linewidth]{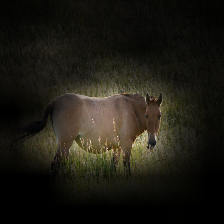}%
   
  \caption{Horse-Body}
  \label{fig:horse_body}
\end{subfigure}
\\
\begin{subfigure}{\textwidth}
  \includegraphics[width=0.12\linewidth]{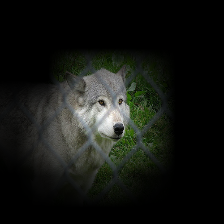}%
  \hfill
  \includegraphics[width=0.12\linewidth]{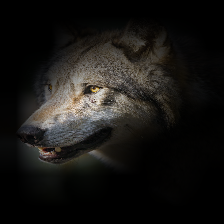}%
    \hfill
  \includegraphics[width=0.12\linewidth]{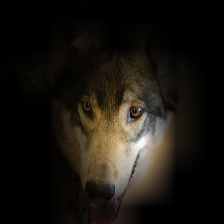}%
    \hfill
  \includegraphics[width=0.12\linewidth]{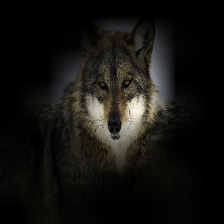}%
  \hfill
  \includegraphics[width=0.12\linewidth]{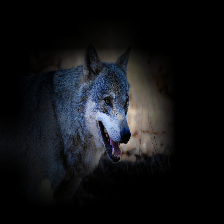}%
  \hfill
  \includegraphics[width=0.12\linewidth]{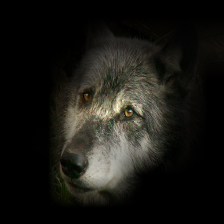}%
   \hfill
  \includegraphics[width=0.12\linewidth]{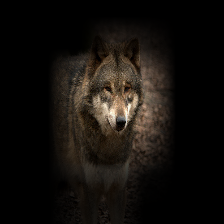}%
   \hfill
  \includegraphics[width=0.12\linewidth]{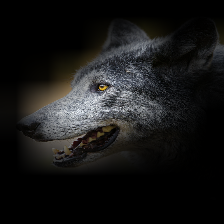}%
  \caption{Wolf-Head}
  \label{fig:wolf_head}
\end{subfigure}
\\
\begin{subfigure}{\textwidth}
  \includegraphics[width=0.12\linewidth]{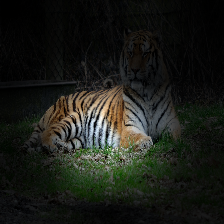}%
  \hfill
   \includegraphics[width=0.12\linewidth]{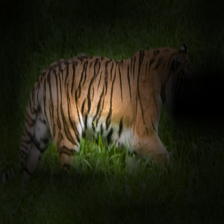}%
    \hfill
  \includegraphics[width=0.12\linewidth]{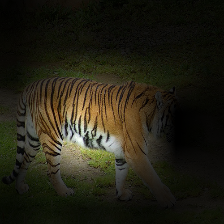}%
    \hfill
  \includegraphics[width=0.12\linewidth]{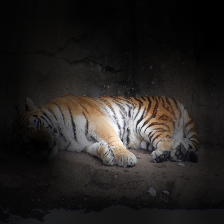}%
  \hfill
  \includegraphics[width=0.12\linewidth]{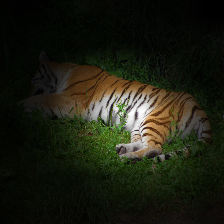}%
  \hfill
  \includegraphics[width=0.12\linewidth]{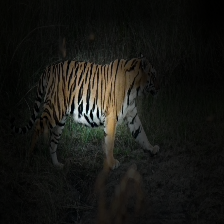}%
  \hfill
  \includegraphics[width=0.12\linewidth]{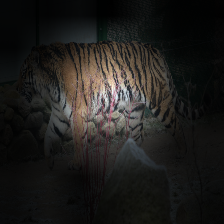}%
  \hfill
  \includegraphics[width=0.12\linewidth]{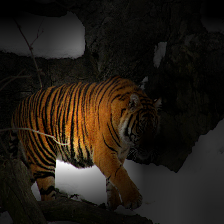}%
  \caption{Tiger-Body}
  \label{fig:tiger_body}
\end{subfigure}
\\
\begin{subfigure}{\textwidth}
  \includegraphics[width=0.12\linewidth]{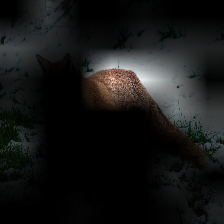}%
  \hfill
   \includegraphics[width=0.12\linewidth]{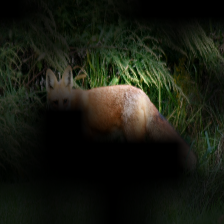}%
    \hfill
   \includegraphics[width=0.12\linewidth]{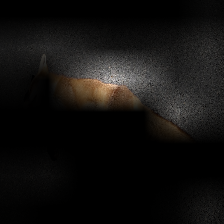}%
    \hfill
   \includegraphics[width=0.12\linewidth]{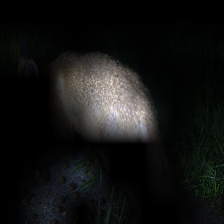}%
  \hfill
   \includegraphics[width=0.12\linewidth]{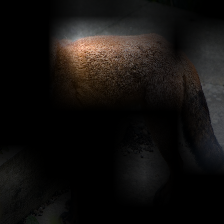}%
  \hfill
   \includegraphics[width=0.12\linewidth]{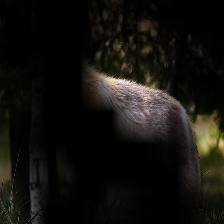}%
  \hfill
   \includegraphics[width=0.12\linewidth]{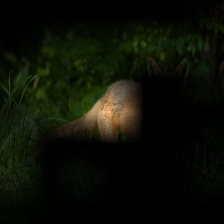}%
  \hfill
   \includegraphics[width=0.12\linewidth]{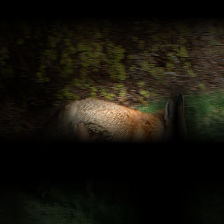}%
  \caption{Fox-Back}
  \label{fig:fox_back}
\end{subfigure}
\\
\begin{subfigure}{\textwidth}
  \includegraphics[width=0.12\linewidth]{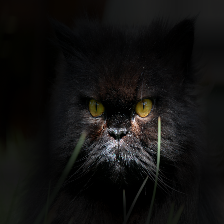}%
  \hfill
   \includegraphics[width=0.12\linewidth]{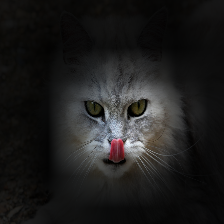}%
    \hfill
   \includegraphics[width=0.12\linewidth]{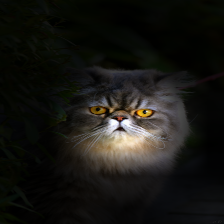}%
    \hfill
   \includegraphics[width=0.12\linewidth]{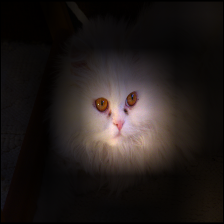}%
  \hfill
   \includegraphics[width=0.12\linewidth]{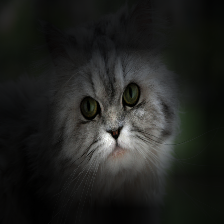}%
  \hfill
   \includegraphics[width=0.12\linewidth]{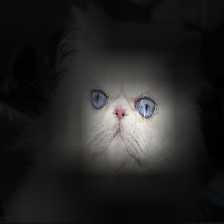}%
  \hfill
   \includegraphics[width=0.12\linewidth]{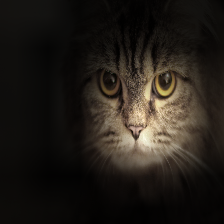}%
  \hfill
   \includegraphics[width=0.12\linewidth]{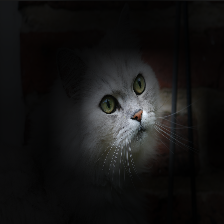}%
  \caption{Cat-Face}
  \label{fig:cat_face}
\end{subfigure}
\caption{ResNet Outputs}
\label{fig:resnet_outputs}
\end{figure}

The visualizations for the ResNet model, trained on AWA2 are shown in Figure \ref{fig:resnet_outputs}. We observed more head and body concepts in the case of the ResNet model, as compared to the VGG model. A possible reason for this could be that ResNet is a much deeper network than the VGG, and hence the features present in the activation maps of the last convolution layer majorly contain high-level body concepts.

\newpage
\section{Ablation Study}

In order to justify the need for both the $\mathcal{L}^O$ and $\mathcal{L}^D$ losses, we compared the faithfulness of the approach with and without these losses. According to our hypothesis, we require the $\mathcal{L}^O$ loss to help us recreate the output of the first dense layer of a classifier, so as to keep our concept embeddings faithful to the classifier. We require $\mathcal{L}^D$ Loss to avoid possibility of inconsistency with the final output of the classifier.

We train two models, one without using the $\mathcal{L}^O$ loss, while other without the $\mathcal{L}^D$ loss. We train both the models for 50 epochs, setting the learning rate to be $10^{-4}$.

\begin{figure}[H]
    \centering
    \includegraphics[width=\textwidth,height=\textheight,keepaspectratio]{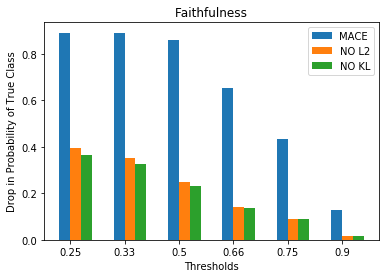}
    \caption{[Best viewed in color] Effect of $\mathcal{L}^O$ and $\mathcal{L}^D$ losses on model faithfulness}
    \label{fig:ablation}
\end{figure}

In Figure \ref{fig:ablation}, we see that there is a significant difference in drop in the true class probability when either of the two losses are reduced, showing both these losses are necessary for the faithfulness of our MACE unit.  

\newpage
\bibliographystyle{plainnat}
\bibliography{neurips2020.bib}